\documentclass{article}

\usepackage{iclr2027_conference,times}

\usepackage{amsmath,amsfonts,bm}

\def\eqref#1{equation~\ref{#1}}

\def\1{\bm{1}}

\DeclareMathAlphabet{\mathsfit}{\encodingdefault}{\sfdefault}{m}{sl}
\SetMathAlphabet{\mathsfit}{bold}{\encodingdefault}{\sfdefault}{bx}{n}

\usepackage[T1]{fontenc}
\usepackage[utf8]{inputenc}
\usepackage{microtype}
\usepackage{inconsolata}
\usepackage{graphicx}
\usepackage{float}
\usepackage{wrapfig}
\usepackage{needspace}
\usepackage{footnote}
\usepackage{subcaption}
\usepackage[inline]{enumitem}
\usepackage[most]{tcolorbox}
\usepackage{adjustbox}
\usepackage{colortbl}
\usepackage{array}
\usepackage{booktabs}
\usepackage{multirow}
\usepackage{xcolor}
\usepackage[strict]{changepage}
\usepackage{framed}
\usepackage{tikz}
\usepackage{hyperref}
\usepackage{url}

\setlist[itemize]{leftmargin=*,itemsep=0pt,topsep=2pt,parsep=0pt,partopsep=0pt}

\definecolor{formalshade}{rgb}{0.95,0.95,1}
\definecolor{DarkPurple}{rgb}{0.5, 0.0, 0.5}
\definecolor{DarkBlue}{rgb}{0.0, 0.5, 0.5}

\newenvironment{formal}{%
\MakeFramed{\advance\hsize-\width\FrameRestore}%
\noindent\hspace{-4.55pt}%
\begin{adjustwidth}{}{-1pt}%
\raggedright%
}
{%
\end{adjustwidth}\endMakeFramed%
}

\definecolor{takeaway_color}{RGB}{238,249,254}

\definecolor{myblue}{RGB}{0,61,115}
\tcbset{
  promptbox/.style={
    enhanced,
    colback=white!5,
    colframe=myblue,
    coltitle=white,
    fonttitle=\bfseries,
    arc=3mm,
    breakable,
    boxrule=1.5pt,
    fonttitle=\bfseries\normalsize,
    title filled,
    colbacktitle=myblue,
    boxed title style={
      sharp corners,
      rounded corners,
      top=15mm,
      minimum height=60mm,
    }
  }
}

\definecolor{myred}{RGB}{107,0,97}
\tcbset{
  reviewbox/.style={
    enhanced,
    colback=white!5,
    colframe=myred,
    coltitle=white,
    fonttitle=\bfseries,
    arc=3mm,
    breakable,
    boxrule=1.5pt,
    fonttitle=\bfseries\normalsize,
    title filled,
    colbacktitle=myred,
    boxed title style={
      sharp corners,
      rounded corners,
      top=15mm,
      bottom=15mm,
      minimum height=10mm,
    }
  }
}

\newcommand{\dashedline}{%
\par\vspace{1mm}%
\tikz \draw[dashed, thick, gray] (0,0) -- (\linewidth,0);%
\vspace{1mm}%
}

\definecolor{darkblue}{rgb}{0, 0, 0.5}
\hypersetup{colorlinks=true, citecolor=darkblue, linkcolor=darkblue, urlcolor=darkblue}

\title{LLM-REVal: Can We Trust LLM Reviewers Yet?}

\author{
\textbf{Rui Li$^1$, Jia-Chen Gu$^2$, Po-Nien Kung$^2$, Heming Xia$^3$, Junfeng Liu$^4$, Xiangwen Kong$^4$,}\\
\textbf{Zhifang Sui$^1$, Nanyun Peng$^2$}\\
$^1$Peking University ~ $^2$University of California, Los Angeles\\
$^3$The Hong Kong Polytechnic University ~ $^4$StepFun AI\\
\texttt{o\_l1ru1@stu.pku.edu.cn}, \texttt{\{gujc, ponienkung\}@ucla.edu}\\
\texttt{heming.xia@connect.polyu.hk}, \texttt{szf@pku.edu.cn}, \texttt{violetpeng@cs.ucla.edu}\\
\href{https://o-l1ru1.github.io/LLM-REVal}{\textbf{\texttt{https://O-L1RU1.github.io/LLM-REVal}}}
}

\iclrfinalcopy
\begin{document}
\maketitle
\lhead{\textsc{LLM-REVal}}
\renewcommand{\headrulewidth}{0.4pt}

\begin{abstract}
The rapid advancement of large language models (LLMs) has encouraged their integration into academic workflows, potentially reshaping how research is conducted and reviewed.
While previous studies highlight the potential of LLMs in supporting research and peer review, their dual roles in academic workflows and the interaction between research and review introduce risks that remain underexplored.
In this study, we examine how the joint use of LLMs in research generation and peer review may affect scholarly fairness, focusing on the risks of using LLMs as reviewers through simulation.
Our simulation incorporates a research agent that generates and revises papers, alongside a review agent that evaluates the submissions.
Based on the simulation results, we conduct human annotations and identify notable misalignment between LLM-based reviews and human judgments:
(1) LLM reviewers assign more favorable scores to LLM-authored papers than to matched human-authored papers;
(2) LLM reviewers persistently underrate human-authored papers on limitation-oriented topics (e.g., risk and fairness), even after multiple rounds of revision. Further analyses investigate these findings through paper-level linguistic profiles, the effect of LLM polishing, and limitation-oriented topics.
These results highlight risks and equity concerns for human authors and scholarly evaluation if LLMs are deployed in peer review without adequate caution.
On the other hand, revisions guided by LLM reviews yield quality improvements, particularly for initially lower-quality papers, illustrating the potential of LLMs as reviewers to support early-stage researchers.
\end{abstract}

\section{Introduction}

Large language models (LLMs) are demonstrating growing autonomy in scientific research, 
spanning tasks such as idea generation~\citep{wang2023scimonscientificinspiration,yang2023largelanguagemodels,qi2023largelanguagemodels,si2024canllmsgenerate,kumar2024canlargelanguage},  automated citation~\citep{press2024citemecanlanguage,ajith2024litsearcharetrieval,kang2024researcharenabenchmarkinglarge}, and data analysis~\citep{huang2023mlagentbenchevaluatinglanguage,tang2023ml-benchevaluatinglarge,guo2024ds-agentautomateddata,tian2024scicodearesearch,chan2024mle-benchevaluatingmachine,app15095208}.
With a substantial increase in submissions and unprecedented pressure on peer review, using LLMs to alleviate review workload~\citep{gao2024reviewer2optimizingreview, d'arcy2024margmulti-agentreview,zhu2025deepreviewimprovingllm-based} has gained interest.
Notably, recent work demonstrated that LLMs can enhance the peer review process by improving the clarity, actionability, and interactivity of reviews \citep{liang2023canlargelanguage,thakkar2025canllmfeedback,ryu2025reviewscoremisinformedpeerreview}, and by generating high-quality meta-review summaries~\citep{hossain2024llmsasmeta-reviewers'}.

Despite their potential, the use of LLMs as reviewers raises profound concerns for the integrity of the scholarly ecosystem~\citep{du2024llmsassistnlp,lin2025breakingreviewerassessingvulnerability, zhu2025your},
as their judgments may introduce differential scoring patterns that compromise evaluative fairness.
Existing studies have identified several vulnerabilities of LLM reviewers, including susceptibility to hidden prompt manipulation and a focus on self-disclosed limitations in manuscripts~\citep{ye2024yetrevealingrisksutilizing,tyser2024ai-drivenreviewsystems, lin2025breakingreviewerassessingvulnerability, zhu2025your}. 
In practice, the concerns are amplified by LLMs' dual roles: they are increasingly used to refine phrasing or even generate manuscripts, while some reviewers, despite explicit prohibitions, delegate their responsibilities to LLMs~\citep{yu2025paperreviewedllmbenchmarking}, creating the prospect that an LLM-refined or even LLM-generated paper may be reviewed by an LLM reviewer.
The systemic implications and potential risks arising from the dual roles and feedback loop of LLMs in the academic process remain largely unexplored.

\begin{figure}[t]
    \centering
    \includegraphics[width=\textwidth]{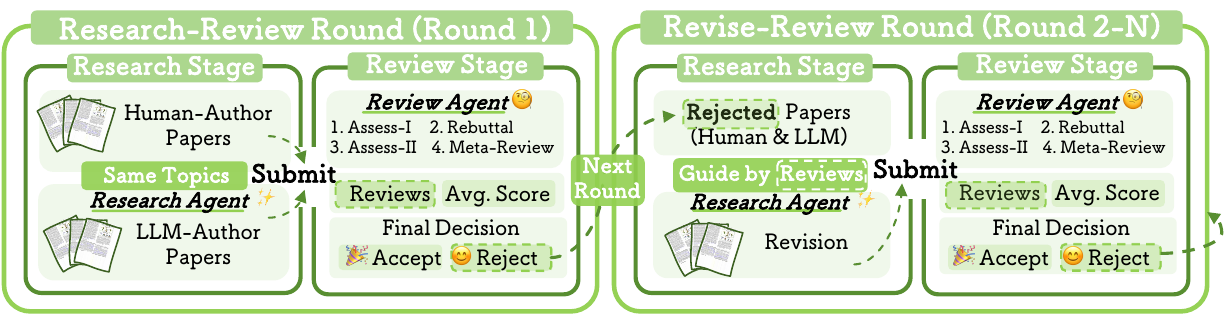}
    \caption{Pipeline and composition of our simulation. 
    In the research-review round, we have human-author papers and LLM-authored papers generated by the research agent as submissions. The review agent then reviews each paper, and the acceptance decision is made based on the review scores. In the revise-review rounds, we take the low-scoring papers from the previous round, revise them guided by LLM reviews, then repeat the same review process as before.}
    \label{fig:simulation-main}
\end{figure}

To address this, we propose \textbf{LLM-REVal} (\textbf{LLM} \textbf{REV}iewer \textbf{R}e-\textbf{EVal}uation) through a multi-round simulation of the academic publication process.
As shown in Figure \ref{fig:simulation-main}, the simulation begins with a research-review round that includes both human-authored and LLM-authored submissions, followed by multiple revise-review rounds where low-scoring papers are resubmitted in revised form. 
The simulation comprises a \textit{Research Agent} and a \textit{Review Agent}.
The \textit{Research Agent} autonomously executes the research workflow, encompassing literature retrieval, idea generation, experimental design, result analysis, manuscript compilation, and subsequent possible revision.  
The generated papers are visually and structurally indistinguishable from human-authored papers. 
The \textit{Review Agent} emulates the complete scholarly peer-review pipeline, including initial reviewer assessments, rebuttals, reviewer reassessments, meta-reviews, and final decisions. The \textit{review agent} exhibits indicative ability for human paper acceptance and correlates well with human review scores, ensuring the reliability of the simulation.

Focusing on potential risks arising from deep integration of LLMs into scholarly workflows, we quantitatively analyze multi-round review results, comparing LLM-authored and human-authored papers, as well as original low-scoring submissions and their revisions. We identify salient patterns in LLM reviewer behavior, contrast these with human judgments, and conduct additional analyses of the factors associated with the observed score differences. Our findings are as follows:

\begin{itemize}
\item Through analyzing LLM-reviewer ratings, we identify \textbf{1)} LLM reviewers \textbf{\textit{assign significantly higher scores to LLM papers}} compared to human papers on the same topic.
\textbf{2)} \textbf{\textit{Revisions show score improvements}} over their low-scoring initial submissions, regardless of authorship.
\textbf{3)} Despite multiple revisions, the LLM reviewer \textbf{\textit{persistently assigns low scores to human-authored papers on limitation-oriented topics (e.g., risk and fairness)}}.
\item The comparisons with human evaluation reveal a misalignment between LLM reviews and human judgments: \textbf{\textit{Human reviewers do not prefer LLM papers over human papers as LLM reviewers do}}, and \textbf{\textit{ those consistently low-scored human papers are deemed valuable.}} This misalignment indicates that LLM-review scores can diverge from expert human judgments.

\item Further analyses reveal: \textbf{\textit{LLM-authored papers exhibit distinct paper-level linguistic profiles}}, including concision, lexical diversity, and complexity; \textbf{\textit{LLM polishing is accompanied by higher review scores and shifts in linguistic features}}; and \textbf{\textit{multiple complementary analyses consistently indicate that LLM reviewers assign relatively lower scores to limitation-oriented papers}}.
\end{itemize}

These findings highlight the risks and fairness concerns posed to human authors with irresponsible deployment of LLMs in peer review, particularly as LLMs become increasingly integrated across multiple stages of the research lifecycle.
Without proper safeguards, the adoption of LLMs as reviewers risks amplifying score disparities and eroding trust in the peer-review process.

\section{Related Work}

\paragraph{AI-assisted Review}
The recent surge in manuscript submissions has placed unprecedented pressure on peer review, motivating growing interest in using LLMs to support the reviewing process. Prior work suggests that LLMs can provide useful feedback on research papers~\citep{liang2023canlargelanguage}; a randomized study of 20,000 ICLR 2025 reviews finds that LLM-assisted feedback improves the clarity, actionability, and interactivity of peer reviews~\citep{thakkar2025canllmfeedback}; and LLMs have also been explored for meta-review summarization~\citep{hossain2024llmsasmeta-reviewers'}. Beyond single-step feedback generation, recent systems develop reviewer agents, including static review-generation systems such as Reviewer2~\citep{gao2024reviewer2optimizingreview}, MARG~\citep{d'arcy2024margmulti-agentreview}, and DeepReview~\citep{zhu2025deepreviewimprovingllm-based}, as well as more interactive or dynamic review systems such as AgentReview~\citep{jin2024agentreviewexploringpeer}, ReviewMT~\citep{tan2024peerreviewmultiturnlongcontext}, and ReviewerToo~\citep{sahu2025reviewertoo}. At the same time, recent studies caution against uncritical deployment: LLM reviewers may overemphasize self-disclosed limitations and assign high scores to submissions with limited substantive content~\citep{ye2024yetrevealingrisksutilizing}, AI-assisted reviewing can affect scores and acceptance rates~\citep{russo2025aireviewlottery}, and automated reviewers may show excessive agreement or be vulnerable to paper laundering~\citep{baumann2026stopautomating}. Other work further reports low sensitivity to ethical concerns and technical nuances~\citep{tyser2024ai-drivenreviewsystems}, vulnerability to adversarial text perturbations~\citep{lin2025breakingreviewerassessingvulnerability}, and susceptibility to hidden prompt manipulations~\citep{ye2024yetrevealingrisksutilizing}.
While these studies reveal important vulnerabilities of LLM-based reviewing, they largely treat reviewing in isolation; our work instead studies LLM reviewers within a coupled research--review simulation where LLMs can participate in both manuscript generation and evaluation.

\paragraph{AI-assisted Research}
LLMs are also increasingly integrated into diverse scientific research workflows, including idea generation~\citep{wang2023scimonscientificinspiration,yang2023largelanguagemodels,qi2023largelanguagemodels,si2024canllmsgenerate,kumar2024canlargelanguage,li2024chainofideas}, automated citation and literature search~\citep{press2024citemecanlanguage,ajith2024litsearcharetrieval,kang2024researcharenabenchmarkinglarge}, writing assistance~\citep{abdulhai2026llmsdistort}, and experiment design, execution, and analysis~\citep{app15095208,huang2023mlagentbenchevaluatinglanguage,tang2023ml-benchevaluatinglarge,tian2024scicodearesearch,chan2024mle-benchevaluatingmachine,guo2024ds-agentautomateddata}. These studies show both the promise and limitations of LLMs as research assistants: LLM-generated ideas can be rated highly for novelty~\citep{si2024canllmsgenerate}, systems such as LitSearch~\citep{ajith2024litsearcharetrieval} and CiteME~\citep{press2024citemecanlanguage} support scholarly retrieval and citation matching, and systems such as MLR-Copilot~\citep{li2024mlr-copilotautonomousmachine} and ResearchAgent~\citep{baek2024researchagentiterativeresearch} automate experiment construction and iterative optimization. More recent end-to-end research pipelines built on commercial LLM APIs, including AI Scientist~\citep{lu2024theaiscientist} and Zochi~\citep{zochi2025}, move from isolated assistance toward autonomous research agents capable of producing complete manuscripts. 
This expanding role of LLMs in research motivates our experimental setting: rather than studying AI-assisted research or AI-assisted review separately, we simulate a scholarly workflow in which LLM-generated or LLM-revised papers are evaluated by LLM reviewers.

\section{Simulation Construction}

Our work simulates the academic publication process through iterative interactions between a \textit{research agent} and a \textit{review agent}. 
The research agent generates ideas, conducts studies, and revises manuscripts, while the review agent evaluates submissions and provides reviews. This multi-round framework enables analysis of LLM-mediated review dynamics and associated risks.

\subsection{Research Agent}
\label{sec:research_agent}

Drawing inspiration from~\cite{lu2024theaiscientist} and~\cite{si2024canllmsgenerate}, we constructed a research agent that initiates the target research project with a list of keywords, which jointly define the research direction.
According to these keywords, the research agent conducts literature retrieval, generates ideas, designs experiments, predicts and analyzes results, drafts the paper, and finally compiles the manuscript from \LaTeX{} to PDF format.

\paragraph{Literature Retrieval}
Retrieval-augmented generation~\citep{gao2024retrievalaugmentedgenerationlargelanguage} with literature is integrated at multiple stages of the research agent workflow to ensure knowledge accuracy.
During idea generation, we query the Semantic Scholar API~\citep{Kinney2023TheSS} with keywords and rank retrieved papers by relevance, empirical grounding, and novelty.
The top-ranked papers are then used to inspire research ideas and guide experimental design.
In the paper writing stage, previously retrieved papers are aggregated into a topic-specific paper bank. We filter this corpus by semantic similarity to select references for drafting.
Additionally, the Google Search API\footnote{\href{hhttps://developers.google.com/custom-search/v1/overview}{https://developers.google.com}} is used to obtain relevant web content, which is summarized to support background sections.

\paragraph{Idea Generation and Experimental Design} 
Based on the retrieved literature and manually curated examples, the research agent generates candidate ideas. 
Cosine similarity is used to remove duplicate ideas, after which the remaining ideas are ranked, and the top idea is selected. 
Building on this idea, the research agent develops an experiment plan and refines it according to demonstrations summarized from the retrieved literature.
Because full experimental execution is costly and difficult to scale with current LLM coding capabilities, we use the research agent to predict experimental results rather than run complete experiment iterations.
\textit{To prevent LLM-authored papers from gaining an artificial advantage through stronger reported results, we cap each predicted numerical improvement over baselines by the corresponding improvement magnitude observed in the matched human-authored paper under the same topic and keyword setting.}

\paragraph{Paper Writing}
We design an effective and efficient workflow that combines rule-based components with LLM-based generation to guide the paper writing process.
\textbf{1) Plug-and-Play Template.} The research agent is initialized with the official ICLR \LaTeX{} template.
Following standard academic structure, the paper is organized into: \textit{Abstract}, \textit{Introduction}, \textit{Background}, \textit{Method}, \textit{Experimental Setup}, \textit{Results Analysis}, \textit{Related Work}, and \textit{Conclusion}.
\textbf{2) Sequential Iteration.} Sections are generated in order, with each step conditioned on all previously generated content and relevant literature, ensuring coherence and consistency. 
\textbf{3) References Integrity.} 
To address the low reference count and citation formatting errors in LLM-authored papers, we implement a citation module. 
The research agent integrates both provided (e.g., retrieved literature) and autonomously generated references (e.g., models, datasets, foundational works) in the writing process. 
For each generated reference, search keywords are extracted from its context and used to locate the original work via Google and arXiv.
References that cannot be reliably verified are removed, ensuring the integrity of the bibliography.
\textbf{4) Incremental Compilation.} After each section, the manuscript is compiled to detect and fix errors early. 
The final manuscript is compiled into a PDF.

The prompt used for paper generation is provided in Appendix \ref{app:prompt_author}, and an example of the generated paper can be found in Appendix \ref{app:example}.

\subsection{Review Agent}
\label{sub:review_agent}
We build our review agent on top of \textsc{AgentReview}~\citep{jin2024agentreviewexploringpeer}. 
Given a paper in PDF format, the system processes it through a five-stage pipeline simulating the peer-review workflow:  
(1) \emph{Reviewer Assessment I},  
(2) \emph{Author--Reviewer Discussion},  
(3) \emph{Reviewer Assessment II},  
(4) \emph{Meta-Review Compilation}, and  
(5) \emph{Final Decision}.

In the \textbf{Reviewer Assessment I} stage, three independent reviewers evaluate the manuscript solely based on its content. Each review includes potential reasons for acceptance and rejection, suggestions for improvement, and a numerical score on a 1--10 scale.
During the \textbf{Author--Reviewer Discussion} stage, simulated authors respond to each review with a rebuttal, addressing misunderstandings, justifying methodological choices, and acknowledging valid critiques.
In the \textbf{Reviewer Assessment II} stage, reviewers revisit their initial evaluations and update both their reviews.
During the \textbf{Meta-Review Compilation} stage, the area chair (AC) synthesizes the reviewers' assessments, producing a meta-review that summarizes the manuscript's strengths and weaknesses, along with a numerical rating.
In the \textbf{Final Decision} stage, we determine acceptance based on the average score across prior stages. Following common practice, we adopt $6$ as the acceptance threshold. Papers with an average score $\geq 6$ are considered accepted. All roles encompassed within the review framework are supported by a single underlying model.

The specific reviews will help us revise the papers in the next round of simulation, while the average scores allow us to identify significant salient scoring patterns in LLM reviewers and further guide the development of multi-round simulations.

\subsection{Simulation Workflow}
\label{sec:simulation_workflow}

As shown in Figure \ref{fig:simulation-main}, the simulation consists of a research-review round and multiple revise-review rounds, each comprising a research stage and a review stage.

\paragraph{Research-Review (Round 1)}
In this round, we prepare submissions from two sources: (i) human-authored papers collected from actual International Conference on Learning Representations (ICLR) submissions, and (ii) LLM-authored papers generated by our research agent.
Details of the human paper collection process are provided in Section~\ref{sec:data_collection}.
Since keywords play a crucial role in shaping the topic and content of LLM-authored papers, we directly used keywords extracted from human-authored papers to guide the papers' generation. 
This design guarantees that LLM-authored papers are aligned with reasonable and feasible research directions, mitigates hallucinations, and enhances the comparability between LLM-authored and human-authored papers.
Subsequently, all submissions are evaluated by the review agent. 
For each paper, we collect the average score and the corresponding detailed reviews. 
The average score serves as the criterion for deciding whether a paper proceeds to the next stage. Papers averaging a score of $\geq$ 6 will be considered \emph{accepted}, other papers will be revised and resubmitted for the next round. 
The detailed feedback guides revisions of papers advancing to the subsequent round.

\paragraph{Revise-Review (Round 2 to Beyond)}
In the revise--review round, we focus on revisions of papers rejected in the previous round, without introducing new submissions. 
For LLM-authored papers, we directly revise the complete \LaTeX{} source generated by the research agent in the previous round, guided by the corresponding LLM reviews. 
For human-authored papers, we first collect the original ICLR \LaTeX{} sources from arXiv and then revise them using the corresponding LLM reviews. 
The rest procedure follows the same process as round 1.
The revise--review cycle will be repeated for up to six rounds, or until 
all previously rejected papers achieve acceptance.

\section{Simulation Statistics and Construction}
\subsection{Models}

We adopted \texttt{DeepSeek-R1} and \texttt{DeepSeek-V3}\footnote{We used the versions \href{https://huggingface.co/deepseek-ai/DeepSeek-R1-0528}{DeepSeek-R1-0528} and \href{https://huggingface.co/deepseek-ai/DeepSeek-V3-0324}{DeepSeek-V3}.} for the \textbf{research agent}.
Following \citet{guo2025deepseek}, \texttt{DeepSeek-R1} exhibits strong reasoning and creativity; we therefore used it for research-idea generation.
Based on our preliminary experiments, \texttt{DeepSeek-V3} was used for the writing stage due to its generation quality, lower hallucination rate.
For the \textbf{review agent}, we evaluate \texttt{GPT-5.5}, \texttt{DeepSeek-V4-Pro}, and \texttt{Gemini-3.1-Pro}\footnote{We used the deployed model identifiers \href{https://learn.microsoft.com/en-us/azure/foundry/openai/how-to/reasoning?tabs=csharp\%2Cgpt-6}{\texttt{gpt-5.5-2026-04-24}}, \href{https://api-docs.deepseek.com/news/news260813/}{\texttt{DeepSeek-V4-Pro}}, and \href{https://docs.cloud.google.com/gemini-enterprise-agent-platform/models/gemini/3-1-pro}{\texttt{Gemini-3.1-Pro}}.} on a common held-out human-submission set, as reported in Section~\ref{sec:review_agent_validation_scope}. We select the best-performing model, \texttt{GPT-5.5}, as the review agent for the end-to-end simulation. We subsequently use the other models to validate the findings drawn from the simulation, confirming that the observed patterns hold across reviewer backbones rather than being specific to a single model.

\vspace{-1mm}
\subsection{Human Paper Collection}
\label{sec:data_collection}

We categorized the papers according to their keywords and identified ten major research topics that collectively defined the primary scope of the submissions in our simulation. These topics spanned LLM research, including reasoning, in-context learning, code generation, evaluation, fine-tuning, and hallucination, as well as continual learning, diffusion-based image generation, transformer attention, and self-supervised learning. For each topic, we randomly sampled 10 papers, resulting in a total of 100 human-authored papers\footnote{Generating a paper required approximately 500k input tokens, and reviewing each paper consumed about 200k input tokens. Driven by both cost and the strong, consistent performance observed in preliminary small-scale experiments, we restricted the initial set of human-authored papers to 100.}. For these selected papers, we retrieved the corresponding ICLR submissions from OpenReview\footnote{\href{https://openreview.net}{https://openreview.net}} and obtained their \LaTeX{} sources and metadata from arXiv\footnote{\href{https://arxiv.org}{https://arxiv.org}}.

\Needspace{18\baselineskip}
\subsection{Review Agent Validation and Scope}
\label{sec:review_agent_validation_scope}

\begin{wraptable}[8]{r}{0.45\textwidth}
\vspace{-10pt}
\centering
\footnotesize
\setlength{\tabcolsep}{2pt}
\renewcommand{\arraystretch}{0.98}
\captionsetup{font= footnotesize,justification=raggedright,singlelinecheck=false,skip=2pt}
\caption{Review agent validation on human-authored submissions.}
\label{tab:review_agent_validation}
\begin{tabular}{@{}lcc@{}}
\toprule
\textbf{Reviewer} & \textbf{Acc.} & \textbf{Pearson's $r$} \\
\midrule
\texttt{GPT-5.5} & 80.0\% & 0.5336 \\
\texttt{DeepSeek-V4-Pro} & 66.0\% & 0.3950 \\
\texttt{Gemini-3.1-Pro} & 78.0\% & 0.5864 \\
\bottomrule
\end{tabular}
\vspace{-10pt}
\end{wraptable}

An effective review framework is a prerequisite for ensuring the reliability of our simulation. To this end, we conducted a preliminary experiment to assess whether the review agent can approximate human-level performance under standard review settings (only human paper involved). Specifically, we examined two aspects: (1) the indication of review agent scores, and (2) the correlation between human review scores and review agent scores. We constructed a balanced dataset of 100 ICLR submissions, comprising 50 rejected papers (official scores 1--4) and 50 accepted papers (official scores 6--9), and applied our review agent to evaluate each paper.

\paragraph{Acceptance Indication.}
We first examined whether the average review score produced by each reviewer backbone could serve as a reliable indicator of acceptance decisions. Using the same average-score threshold of $6$ as in our simulated decision rule, papers with an average score $\geq 6$ were predicted as accepted, whereas those below the threshold were predicted as rejected. On the common validation set, \texttt{GPT-5.5} achieved an accuracy of \textbf{80.0\%}, \texttt{DeepSeek-V4-Pro} achieved \textbf{66.0\%}, and \texttt{Gemini-3.1-Pro} achieved \textbf{78.0\%}, indicating a moderate alignment between the predictions and the actual acceptance decisions.

\paragraph{Score Correlation.}
We further evaluated the correspondence between LLM review scores and human review scores. Pearson's correlation coefficients were $r=\mathbf{0.5336}$ for \texttt{GPT-5.5}, $r=\mathbf{0.3950}$ for \texttt{DeepSeek-V4-Pro}, and $r=\mathbf{0.5864}$ for \texttt{Gemini-3.1-Pro}. These results indicate positive associations between LLM and human review scores across all three reviewer backbones. This level of agreement is consistent with prior evidence that LLM--human review overlap can be comparable to inter-human reviewer overlap \citep{thakkar2025canllmfeedback}.

\section{What Salient Patterns Do LLM Reviewers Emerge?}
\label{sec:salient_patterns}
\label{sec:human_check}

From our simulation results, we observed three salient patterns: \textit{LLM-Authored Superiority}, \textit{Revision Boost}, and \textit{Inevitable Rejection}. For each pattern, we report a corresponding human check to examine whether the LLM-review behavior aligns with human judgments.

\Needspace{18\baselineskip}
\subsection{Research-Review (Round 1)}

\vspace{-1mm}
\label{sec:round_1}
\begin{formal}
    \textbf{Pattern 1: Higher Scores for LLM-Authored Papers.} LLM-authored papers achieved significantly higher LLM-review scores than human-authored papers.
\end{formal}

\begin{wraptable}[12]{r}{0.6\textwidth}
    \vspace{-8pt}
    \centering
    \footnotesize
    \setlength{\tabcolsep}{2pt}
    \renewcommand{\arraystretch}{0.98}
    \captionsetup{font=footnotesize,justification=raggedright,singlelinecheck=false,skip=2pt}
    \caption{Comparison of human-authored and LLM-authored papers in terms of average review score, head-to-head win rate, and acceptance rate across reviewer backbones.}
    \label{tab:hp_vs_lp_review_score}
    \begin{tabular}{@{}llccc@{}}
        \toprule
        \textbf{Reviewer} & \textbf{Paper Type}  & \textbf{Avg Score} & \textbf{Win Rate} & \textbf{Acc Rate} \\
        \midrule
        \multirow{2}{*}{\texttt{GPT-5.5}} & Human & 5.4367 & 19\% & 47\% \\
        & LLM & 5.7700 & 47\% & 56\% \\
        \midrule
        \multirow{2}{*}{\texttt{DeepSeek-V4-Pro}} & Human & 5.6867 & 19\% & 49\% \\
        & LLM & 5.8600 & 42\% & 71\% \\
        \midrule
        \multirow{2}{*}{\texttt{Gemini-3.1-Pro}} & Human & 5.6533 & 29\% & 47\% \\
        & LLM & 5.7900 & 43\% & 71\% \\
        \bottomrule
    \end{tabular}
\vspace{-10pt}
\end{wraptable}

As shown in Table~\ref{tab:hp_vs_lp_review_score}, LLM-authored papers received higher scores, pairwise win rates, and acceptance rates than matched human-authored papers across all three reviewer backbones. The average-score advantage was +0.14 to +0.33 points, accompanied by 14--28 percentage-point higher win rates and 9--24 percentage-point higher acceptance rates.

As a sampling sanity check, the set of human-authored submissions used in the simulation was not constructed by oversampling weak or rejected papers. It covers 10 topics with 10 papers each, comprising 41 accepted papers with an average OpenReview score of 6.46 and 59 rejected papers with an average OpenReview score of 4.68. Its 41\% acceptance rate substantially exceeds the conference-wide acceptance rate.

\paragraph{Human check.}
We selected the 15 matched human--LLM paper pairs with the largest LLM-favoring score gaps, where LLM-authored papers scored about 0.8 points higher on average. 
These pairs covered six research topics: evaluation, reasoning, in-context learning, hallucination, code generation, and fine-tuning.
Thirteen annotators with graduate-level or higher training in computer science evaluated the pairs using ICLR-style review criteria, without being informed of the papers' authorship.
Each pair was independently judged by two annotators, who selected the stronger paper or a ``tie.'' Annotators selected human-authored papers as superior in \textbf{56.7}\% of cases, compared to \textbf{33.3}\% for LLM-authored papers, indicating a misalignment between LLM and human reviewers. 
This discrepancy raises a practical concern: differences in the degree of LLM involvement in research may introduce reviewer bias and compromise the fairness of peer review.

\subsection{Revise-Review (Round 2 to 6)}

\vspace{-1mm}
\label{sec:round_2}
\begin{formal}
    \textbf{Pattern 2: Revision Boost.} Paper revisions showed improvements in review scores.
\end{formal}

\begin{wraptable}[9]{r}{0.5\textwidth}
    \vspace{-8pt}
    \centering
    \footnotesize
    \setlength{\tabcolsep}{3pt}
    \renewcommand{\arraystretch}{1.0}
    \captionsetup{font=footnotesize,justification=raggedright,singlelinecheck=false,skip=2pt}
    \caption{Average paired scores before and after revision.}
    \label{tab:revision_avg_score}
    \scalebox{0.9025}{%
    \begin{tabular}{@{}lrrrr@{}}
        \toprule
        \multirow{2}{*}{\textbf{Reviewer Model}} & \multicolumn{2}{c}{\textbf{Human Paper}} & \multicolumn{2}{c}{\textbf{LLM Paper}} \\
        \cmidrule(lr){2-3}\cmidrule(lr){4-5}
        & \textbf{Before} & \textbf{After} & \textbf{Before} & \textbf{After} \\
        \midrule
        \texttt{GPT-5.5} & 4.992 & 5.294 & 5.373 & 5.799 \\
        \texttt{DeepSeek V4 Pro} & 5.207 & 5.471 & 5.725 & 6.000 \\
        \texttt{Gemini-3.1 Pro} & 5.771 & 6.036 & 5.649 & 5.921 \\
        \bottomrule
    \end{tabular}%
    }
\end{wraptable}

\paragraph{Round 2}
For human-authored papers, after the initial simulation we selected submissions that received scores below 6 from all three reviewer models for revision and re-review. Because the original \LaTeX{} sources on arXiv were unavailable for some papers, we included only those with accessible original \LaTeX{} sources and content that closely matched their ICLR submission versions.

\paragraph{Human check.}

Across 22 pairs of original papers and LLM-guided revisions, we manually assessed whether issues raised in the original reviews were addressed in the revised papers. To focus the annotation, we used \texttt{GPT-5} to classify review comments into two groups: overlapping comments, which appeared in reviews of both the original and revised papers, and original-only comments, which no longer appeared in reviews of the revised papers. We then mapped each original-only issue to the corresponding paragraphs in both versions, and two annotators independently assessed whether it had been addressed by the revision. After excluding 56 issues that could not be addressed through textual revisions, such as requests for new visualizations, models, or baselines, 81 of the remaining 174 issues (\textbf{46.55\%}) were addressed. This pattern is consistent with the higher scores assigned by the LLM reviewer.

\vspace{-1mm}
\label{sec:round_3}
\begin{formal}
    \textbf{Pattern 3: Inevitable Rejection.} After multiple revise--review cycles, certain types of initial human-authored submissions remained unaccepted.
\end{formal}

Among the human-authored papers that continued to receive relatively low scores even after Round 6, we found that most were limitation- or risk-oriented studies. These papers included diagnostic analyses of model failures, bias and fairness audits, and jailbreak-attack studies.

\begin{center}
\footnotesize
\newcommand{\revisiontablescale}{0.5\linewidth} %

\captionof{table}{Mean scores across revision rounds for limitation-oriented or risk-oriented human papers.}
\label{tab:gpt55_deepseek_strict_revision_scores}
\resizebox{\revisiontablescale}{!}{%
\begin{tabular}{lcccccc}
\toprule
\textbf{Reviewer} & \multicolumn{6}{c}{\textbf{Mean score by revision round}} \\
\cmidrule(lr){2-7}
& \textbf{v0} & \textbf{v1} & \textbf{v2} & \textbf{v3} & \textbf{v4} & \textbf{v5} \\
\midrule
\texttt{GPT-5.5} & 5.00 & 4.85 & 4.75 & 4.85 & 4.80 & 4.75 \\
\texttt{DeepSeek-V4-Pro} & 5.60 & 5.80 & 5.70 & 5.40 & 5.60 & 5.15 \\
\texttt{Gemini-3.1-Pro} & 5.70 & 5.70 & 5.80 & 6.05 & 6.10 & 5.75\\
\bottomrule
\end{tabular}%
}
\end{center}

\vspace{-3mm}
\paragraph{Human check.}
We examined the corresponding OpenReview records and found mean review scores ranging from 4.5 to 6.5, with several papers accepted at ICLR. These results suggest that LLM reviewers may be more likely to assign lower scores to papers whose primary contribution is exposing failures, risks, or limitations, even when their expert-assigned scores are comparable to those of method-oriented controls.

\section{Additional Analysis}

To further investigate these findings, we examine three related aspects: differences in linguistic profiles between LLM- and human-authored papers, the effects of LLM polishing on LLM-review scores, and LLM reviewers' tendency to assign relatively lower scores to limitation-oriented papers.

\subsection{Linguistic Profile of LLM-Authored Papers}
\label{sec:linguistic_feature}

\begin{figure}[!t]
    \centering
    \begin{subfigure}[b]{0.495\textwidth}
        \centering
        \includegraphics[width=\linewidth]{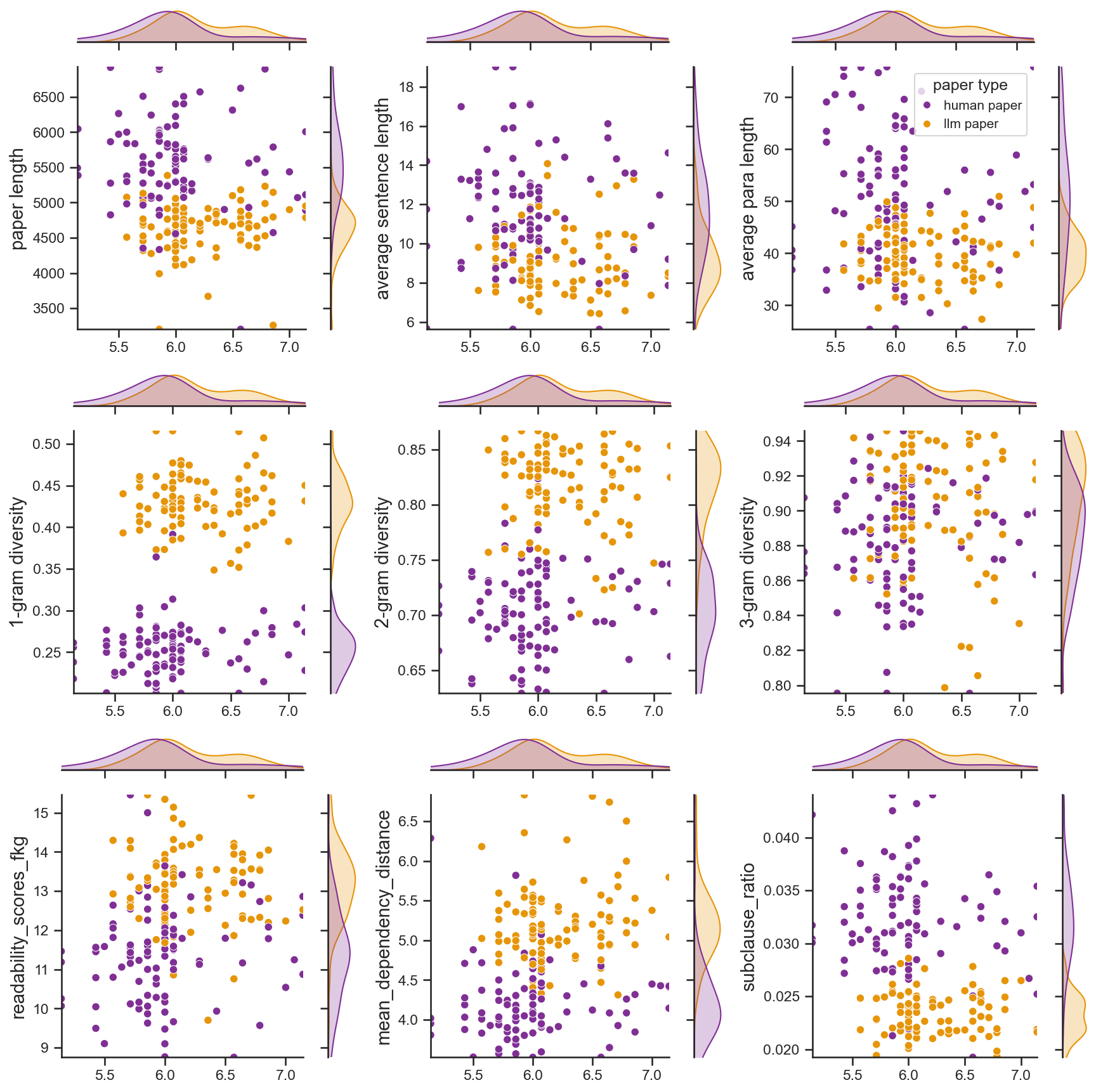}
        \caption{LLM and human papers.}
        \label{fig:llm_human_sta}
    \end{subfigure}
    \hfill
    \begin{subfigure}[b]{0.495\textwidth}
        \centering
        \includegraphics[width=\linewidth]{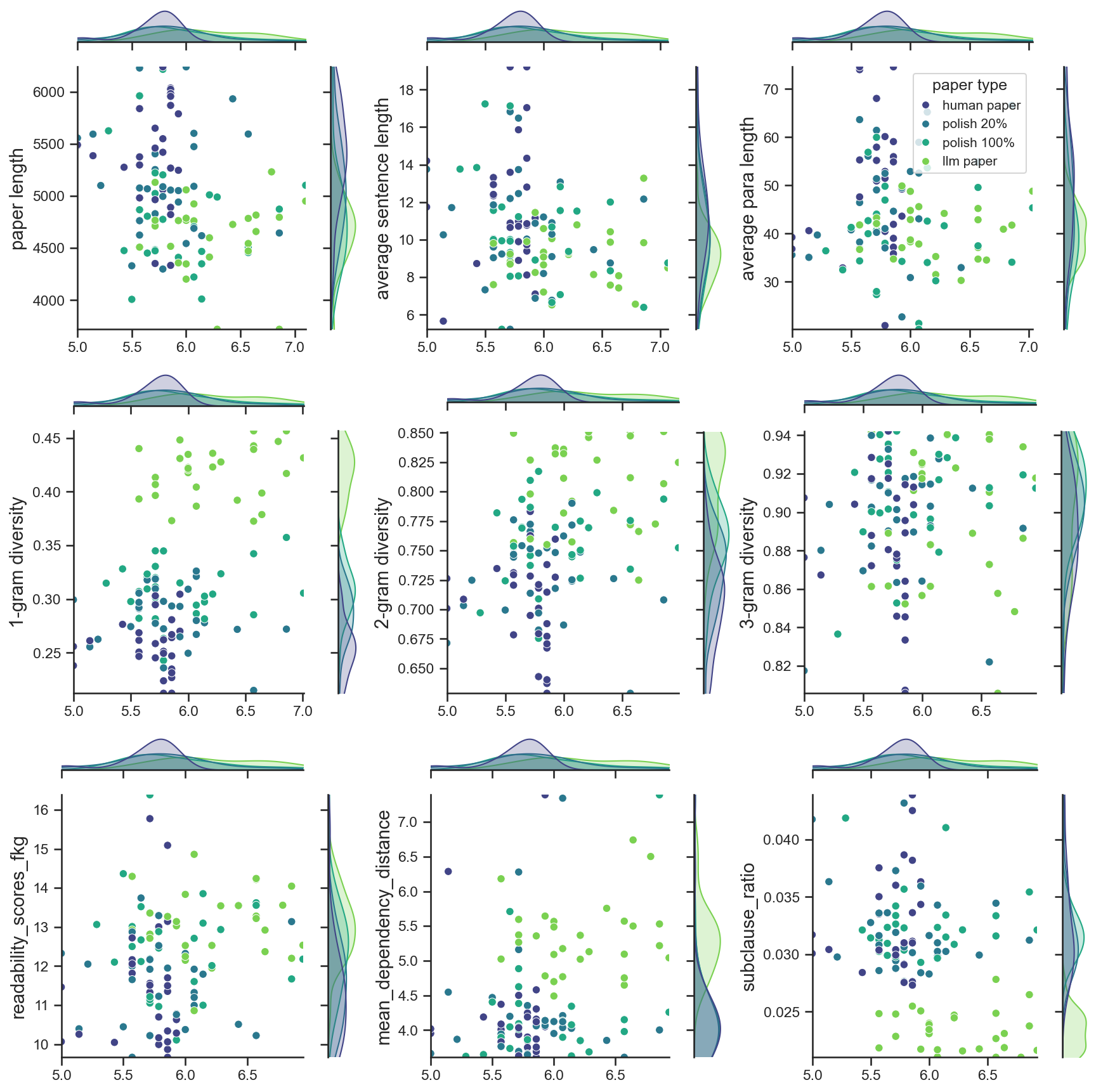}
        \caption{Human, polished, and LLM papers.}
        \label{fig:human_polish_ratio_sta}
    \end{subfigure}
    \caption{Linguistic feature statistics and review score distributions for papers. Each subplot presents a scatter plot with marginal kernel density plots showing the distributions of a specific linguistic feature metric. The x-axis shows the review score, and the y-axis shows different metrics.}
    \label{fig:two_sta}
\end{figure}

We compare the linguistic features of LLM-authored and human-authored papers and identify clear distributional differences along three dimensions: \textbf{length}, \textbf{lexical diversity}, and \textbf{complexity}. Figure~\ref{fig:llm_human_sta} presents these differences, and Appendix~\ref{app:additional_stats} reports the detailed statistical results. These dimensions capture observable differences in writing style.

LLM-authored papers are generally more \textbf{concise}, with shorter overall paper, sentence, and paragraph lengths. This contrasts with prior findings that LLMs tend to favor longer responses~\citep{cai2025disentanglinglengthbiaspreference}.
LLM-authored papers also exhibit higher \textbf{lexical diversity}, measured as the proportion of distinct $n$-grams for $n\in\{1,2,3\}$: 1-gram diversity is 0.4321 for LLM-authored papers, compared with 0.2598 for human-authored papers. They also exhibit distinct \textbf{complexity patterns}: higher lexical complexity, as measured by the lesch--Kincaid Grade Level, and syntactic differences characterized by greater dependency distance but a lower subclause ratio. Together, these patterns suggest longer-range syntactic connections within a shallower hierarchical structure.

\subsection{Effect of LLM Polishing}

We further examine LLM-polished versions of human-authored papers as a descriptive check. The LLM rewrites papers paragraph by paragraph while preserving the original content, such as the reported results. Review scores increased significantly with the polishing ratio. Specifically, human papers with an original average score of 5.69 reached an average score of 5.94 after 40\% LLM polishing, potentially transforming a previously rejected paper into an accepted one. As shown in Figure~\ref{fig:human_polish_ratio_sta} and Figure~\ref{app:app_polish_ratio_sta}, polishing also shifted several aggregate linguistic features statistics toward those of LLM-authored papers.

\subsection{Limitation-Oriented Topics}
\label{sec：critical_statement}
\label{sec:pos_neg}

\begin{wraptable}[9]{r}{0.4\textwidth}
\vspace{-6pt}
\centering
\footnotesize
\setlength{\tabcolsep}{3pt}
\renewcommand{\arraystretch}{0.98}
\captionsetup{font=footnotesize,justification=raggedright,singlelinecheck=false,skip=2pt}
\caption{Average score gaps between limitation-oriented papers and matched method-oriented controls. Negative values favor the controls.}
\label{tab:limitation_control_score_gap}
\begin{tabular}{@{}lc@{}}
\toprule
\textbf{Reviewer} & \shortstack{\textbf{Score gap}} \\
\midrule
\texttt{GPT-5.5} & $-0.475$ \\
\texttt{DeepSeek-V4-Pro} & $-0.313$ \\
\texttt{Gemini-3.1-Pro} & $-0.475$ \\
\bottomrule
\end{tabular}
\end{wraptable}
\vspace{2pt}

In Section~\ref{sec:human_check}, we observed that human-authored papers focused on diagnosing or stress-testing model limitations, rather than proposing new capability-oriented methods, tended to receive less favorable evaluations from LLM reviewers. These papers addressed topics including fairness risks, adversarial attacks, and analyses of model limitations. 
To reduce confounding from paper quality, we paired each limitation-oriented paper with a method-oriented control paper (e.g., one proposing a new model, training method, or optimization algorithm) with the same real human-review score.
Across three LLM reviewer backbones, limitation-oriented papers received lower simulated scores than their matched controls. Specifically, the average score gap (limitation-oriented minus control) was $-0.475$ for \texttt{GPT-5.5}, $-0.313$ for \texttt{DeepSeek-V4-Pro}, and $-0.475$ for \texttt{Gemini-3.1-Pro}. These consistent gaps suggest that LLM reviewers may systematically undervalue limitation-oriented research.

\begin{figure}[!t]
    \centering
    \begin{subfigure}[b]{0.49\textwidth}
        \centering
        \includegraphics[width=\linewidth]{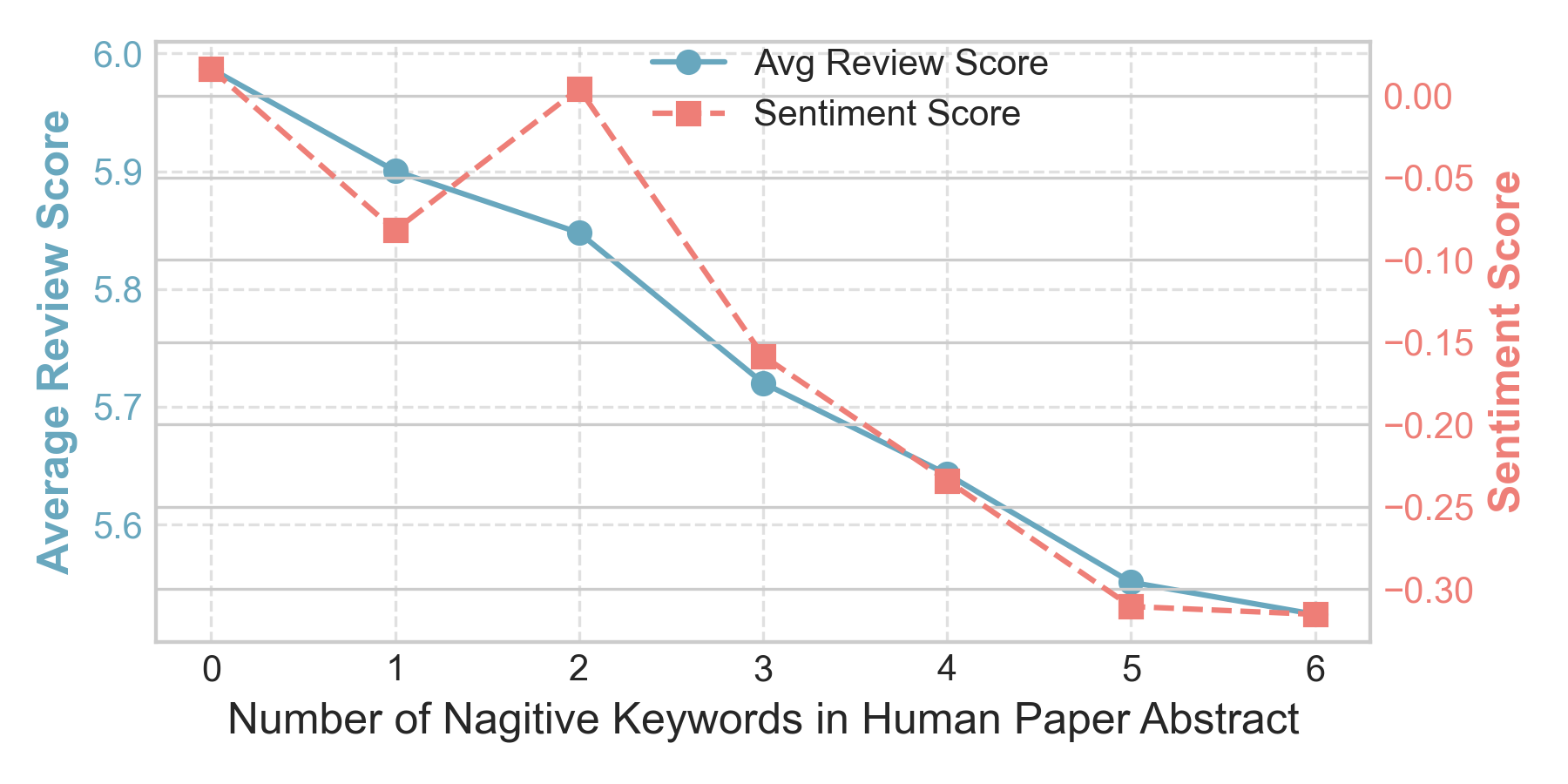}
        \label{fig:negative_score_review_score_human}
    \end{subfigure}
    \hfill
    \begin{subfigure}[b]{0.49\textwidth}
        \centering
        \includegraphics[width=\linewidth]{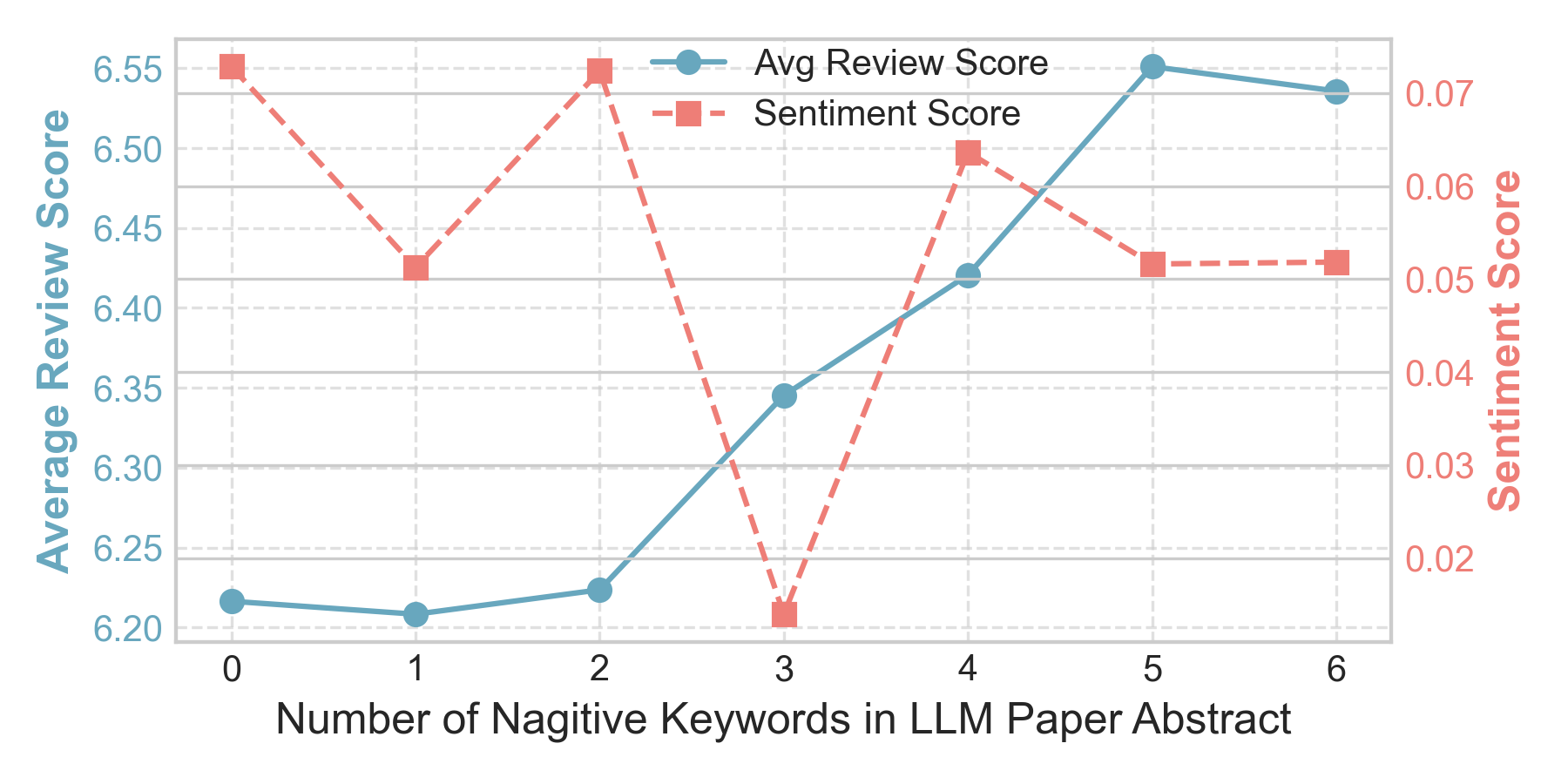}
        \label{fig:negative_score_review_score_llm}
    \end{subfigure}
    \vspace{-0.3cm}
    \caption{Correlation between the number of negative keywords in the abstract, average review score, and sentiment score. The plot shows the trends of both average review score and sentiment score as the number of negative keywords increases.}
    \label{fig:negative_score_review_score}
\end{figure}

Using keyword-based detection, we identified these papers and computed the sentiment polarity of their abstracts; the full keyword list is provided in Table~\ref{tab:negative_keywords}. As shown in Figure~\ref{fig:negative_score_review_score}, among human-authored papers, the frequency of negative keywords was positively correlated with sentiment polarity but negatively correlated with review scores. In contrast, for LLM-authored papers, sentiment polarity showed no clear trend and remained consistently positive regardless of the number of negative keywords, while review scores increased as such keywords became more frequent. This suggests that negatively framed (low-sentiment) discussions of critical topics may exacerbate bias in LLM reviews, whereas positively framed (high-sentiment) discussions of similar topics may receive disproportionately high scores.

We speculate that LLMs' tendency to assign lower scores to limitation-oriented papers related to alignment processes of LLM that favor more positive responses. Consequently, papers that foreground risks, harms, and biases may be penalized by LLM reviewers despite their real-world value.

\section{Conclusion}
Our multi-round simulation of LLM-driven research and review reveals notable review-score disparities when LLMs serve as reviewers. LLM reviewers assign more favorable scores to LLM-authored papers and revised submissions, while some human-authored work on limitation-oriented topics receives less favorable evaluations, leading to misalignment with human judgments. These disparities are associated with LLM-like linguistic profiles and limitation-oriented topics. Our findings highlight that LLM reviewers should not be treated as stand-alone substitutes for expert human reviewers, especially when evaluating LLM-generated research; careful auditing is essential for future AI-integrated scientific workflows.

\section*{Ethics Statement}
This study examines fairness risks that may arise when large language models (LLMs) are used as reviewers in scholarly publishing. All human-authored papers in our simulation were publicly available via OpenReview and arXiv and processed according to their terms of use; no private or non-public reviewer data were accessed. The simulation was conducted offline and did not influence real peer review decisions. Our findings underscore the need for transparency, audits for differential scoring, human oversight, and rigorous ethical guidelines before deploying LLM-based reviewers in real-world settings.

\section*{Reproducibility Statement}
We support reproducibility by documenting all simulation components, including the research-agent and review-agent pipelines, publicly accessible APIs, and model versions. The Appendix provides prompts for paper generation, revision, and review, as well as examples of generated manuscripts. Human-authored papers are sampled from publicly available ICLR submissions on OpenReview, with corresponding \LaTeX{} sources obtained from arXiv. Sections 3 and 4 describe the full simulation workflow, and the statistical analyses, feature extraction procedures, evaluation metrics, correlation computations, and hypothesis tests are specified for reproduction with standard NLP and statistical toolkits.

\section*{AI Use Statement}
Large Language Models (LLMs) were used to aid in the writing and polishing of the manuscript. Specifically, we used an LLM to assist in refining the language, improving readability, and ensuring clarity in various sections of the paper. The model helped with tasks such as sentence rephrasing, grammar checking, and enhancing the overall flow of the text. The authors take full responsibility for the content of the manuscript, including any text generated or polished by the LLM. We have ensured that the LLM-generated text adheres to ethical guidelines and does not contribute to plagiarism or scientific misconduct.

\bibliography{llm_reval_iclr2027}
\bibliographystyle{iclr2027_conference}

\appendix
\section*{Appendix}
\section{LLM Usage}

Large Language Models (LLMs) were used to aid in the writing and polishing of the manuscript. Specifically, we used an LLM to assist in refining the language, improving readability, and ensuring clarity in various sections of the paper. The model helped with tasks such as sentence rephrasing, grammar checking, and enhancing the overall flow of the text. The authors take full responsibility for the content of the manuscript, including any text generated or polished by the LLM.  We have ensured that the LLM-generated text adheres to ethical guidelines and does not contribute to plagiarism or scientific misconduct.

\section{Prompt}
\subsection{Prompt for Author Agent}
\label{app:prompt_author}
The prompts related to idea generation are consistent with \cite{si2024canllmsgenerate}, while the remaining prompts employed for paper generation are as follows.

\begin{tcolorbox}[
    promptbox,
    title=Abstract,
]
Please generate a paper's abstract based on the provided information.

Tips:

- TL;DR of the paper

- What are we trying to do and why is it relevant?

- Why is this hard?

- How do we solve it (i.e., our contribution!)

- How do we verify that we solved it (e.g., Experiments and results)

- Please make sure the abstract reads smoothly and is well-motivated. This should be one continuous paragraph with no breaks between the lines.

- Just wrap it with \\begin{{abstract}} and \\end{{abstract}}, and the content should be in standard LaTeX language.

- Please ensure it's ICLR-style academic writing, and wrap LaTeX content in \verb|```latex```|

\#\#\#\# Paper Information Begin \#\#\#\#

\{paper\_information\}

\#\#\#\# Paper Information End \#\#\#\#

\#\#\#\# related work abstract Begin \#\#\#\#

\{other\_abstracts\}

\#\#\#\# related work abstract End \#\#\#\#

\end{tcolorbox}

\begin{tcolorbox}[
    promptbox,
    title=Introduction,
]
Generate the paper introduction:

Tips:

- Longer version of the Abstract, i.e. of the entire paper

- What are we trying to do and why is it relevant?

- Why is this hard? 

- How do we solve it (i.e. our contribution!)

- How do we verify that we solved it (e.g. Experiments and results)

- New trend: specifically list your contributions as bullet points

- Extra space? Future work!

- Natural paragraphs are a more standard way of expression than lists. Try not to use lists. You can consider using list when summarizing contributions.

- Using the correct citation format. et al. Should not appear directly in the text. Instead, the citation format \textbackslash cite, \textbackslash citet, \textbackslash citep should be used correctly.

- Please ensure it's ICLR-style academic writing, and wrap LaTeX content in \verb|```latex```|
.

\#\#\#\# Abstract Begin \#\#\#\#

\{abstract\}

\#\#\#\# Abstract End \#\#\#\#

\#\#\#\# paper information Begin \#\#\#\#

\{paper\_information\}

\#\#\#\# paper information End \#\#\#\#

\#\#\#\# part related work for citation Begin \#\#\#\#

\{related\_work\}

\#\#\#\# part related work for citation End \#\#\#\#

\end{tcolorbox}

\begin{tcolorbox}[
    promptbox,
    title={Related Work}
]
Generate the related work section:

Tips:

- Academic siblings of our work, i.e. alternative attempts in literature at trying to solve the same problem.

- Goal is to ``Compare and contrast'' - how does their approach differ in either assumptions or method? If their method is applicable to our Problem Setting I expect a comparison in the experimental section. If not, there needs to be a clear statement why a given method is not applicable.

- The general format is 2 to 4 paragraphs. Each paragraph starts with a summary (the category of works) in bold typeface, followed by an introduction to the relevant works as many as much.

- Using the correct citation format. et al. Should not appear directly in the text. Instead, the citation format \textbackslash cite,\textbackslash citet,\textbackslash citep should be used correctly.

- Please ensure it's latex-style academic writing, and wrap academic writing content in \verb|```latex```|.

\#\#\#\# Available References and Their Information BEGIN \#\#\#\#

\{reference\_datas\}

\#\#\#\# Available References and Their Information end \#\#\#\#

\end{tcolorbox}

\begin{tcolorbox}[
    promptbox,
    title=Background,
]
Please give me a paper background outline based on the following introduction, containing 3\-4 subsections, and finally provide me with 4\-6 keywords for retrieval augmentation (return to me in json format)
\verb|```json{{keywords:[]}}```|

\#\#\#\# Introduction Begin \#\#\#\#

\{introduction\}

\#\#\#\# Introduction End \#\#\#\#

\#\#\#\# paper information Begin \#\#\#\#

\{paper\_information\}

\#\#\#\# paper information End \#\#\#\#

\dashedline

Determine whether the content of the following webpage is relevant to \{query\}. If it is relevant, please extract the theoretical background from the webpage. If it is not relevant, simply return None without outputting any content.

\#\#\#\# webpage content Begin \#\#\#\#

\{text\}

\#\#\#\# webpage content End \#\#\#\#

\dashedline

Generate the background SECTION in \verb|```latex```| according to the background outline and combine it with the retrieved information. with the following requirements:

Tips:

- Academic Ancestors of our work, i.e. all concepts and prior work that are required for understanding our method. 

- Note: If our paper introduces a novel problem setting as part of its contributions, it is best to have a separate Section.

- Appropriately incorporate variables/formulas/formal expressions

- Please ensure it is ICLR-style academic writing, and wrap LaTeX content in \verb|```latex```|.

- Do not use any list formatting, use paragraph format instead.

- Please strictly follow the outline, expand and enrich the outline into the background section.

- All cite should be in bibtex format. Using the correct citation format. The phrase ``et al.'' should not appear directly in the text. Instead, the citation format \textbackslash cite, \textbackslash citet, \textbackslash citep should be used correctly.

\#\#\#\# background outline Begin \#\#\#\#

\{json\_content\_real\}

\#\#\#\# background outline End \#\#\#\#

\#\#\#\# retrieved information Begin \#\#\#\#

\{ex\_texts\}

\#\#\#\# retrieved information End \#\#\#\#
    
\end{tcolorbox}

\begin{tcolorbox}[
    promptbox,
    title=Method,
]

Generate the methology section in latex format, with the following requirements:

Tips:

- What we do. Why we do it. All described using the general Formalism introduced in the Problem Setting and building on top of the concepts / foundations introduced in Background.

- Appropriately incorporate variables/formulas/formal expressions

- Please ensure it's ICLR-style academic writing, and wrap LaTeX content in \verb|```latex```|.

- Do not use any list formatting, use paragraph format instead.

- Be sure to pay attention to latex and avoid any formatting errors

- Using the correct citation format. et al. Should not appear directly in the text. Instead, the citation format \textbackslash cite, \textbackslash citet,\textbackslash citep should be used correctly.

\#\#\#\# Introduction Begin \#\#\#\#

\{introduction\}

\#\#\#\# Introduction End \#\#\#\#

\#\#\#\# Background Begin \#\#\#\#

\{background\}

\#\#\#\# Background End \#\#\#\#

\#\#\#\# paper\_information Begin\#\#\#\#

\{paper\_information\}

\#\#\#\# paper\_information End \#\#\#\#

\end{tcolorbox}

\begin{tcolorbox}[
    promptbox,
    title=Experiment Setting,
]

Based on the completed sections of the paper and the experimental data and its introduction, generate an outline for the ``Experiment Setting'' section.

\#\#\#\# Completed sections of the paper BEGIN \#\#\#\#

\{paper\_already\}

\#\#\#\# Completed sections of the paper END \#\#\#\#

\#\#\#\# Experimental results and their introduction BEGIN \#\#\#\#

\{experiment\_results\_and\_introduction\}

\#\#\#\# Experimental results and their introduction END \#\#\#\#

\dashedline

Based on the writing outline, the already completed parts of the paper, the experimental data, and their introduction, complete the ``Experiment Setting'' section in the style of ICLR academic writing.

- Paragraphs are a more formal way of expression than lists, so try to use paragraphs instead of lists.
- Do not reference non-existent content, such as figures that do not exist at all.

- For datasets, baselines, and open-source models, try to add citations. For closed-source models, you can add links.

- Use the correct citation format. ``et al.'' should not appear directly in the text. Instead, the citation formats \textbackslash cite, \textbackslash citet, \textbackslash citep should be used correctly.

- Please ensure it's ICLR-style academic writing, and wrap LaTeX content in \verb|```latex```|.

\#\#\#\# Writing Outline Begin \#\#\#\#

\{outline\_for\_setting\}

\#\#\#\# Writing Outline End \#\#\#\#

\#\#\#\# Already Completed Paper Part Begin \#\#\#\#

\{paper\_already\}

\#\#\#\# Already Completed Paper Part End \#\#\#\#

\#\#\#\# Experiment Results Begin \#\#\#\#

\{experiment\_results\_and\_introduction\}

\#\#\#\# Experiment Results End \#\#\#\#

\end{tcolorbox}

\begin{tcolorbox}[
    promptbox,
    title=Result Analysis,
]

Based on the completed sections of the paper, experimental setup, results data, and their introduction, generate an outline for the results chapter.

\#\#\#\# Already Completed Paper Section Begin \#\#\#\#
\{paper\_already\}
\#\#\#\# Already Completed Paper Section End \#\#\#\#

\#\#\#\# Experimental Setup Begin \#\#\#\#
\{experiments\_setting\}
\#\#\#\# Experimental Setup End \#\#\#\#

\#\#\#\# Experimental Results and Introduction Begin \#\#\#\#
\{experiment\_results\_and\_introduction\}
\#\#\#\# Experimental Results and Introduction End \#\#\#\#

\dashedline

Based on the writing outline, the completed parts of the paper, and the experimental data along with their introduction, complete the results section in the style of ICLR academic writing.

Tips:

- Paragraphs are a more formal way of expression than lists; try to use paragraphs instead of lists.

- Do not reference non-existent content, such as figures that do not exist at all.

- All Tables should be in LaTeX format 

- All Tables should be included in your generated results section.

- All tables must be with detailed analysis and interpretation of the data.

- Do not modify the content of the table and do not add any note in the table. 

- Please ensure it's ICLR-style academic writing, and wrap LaTeX content in \verb|```latex```|.

\#\#\#\# Writing Outline Begin \#\#\#\#

\{outline\_for\_results\}

\#\#\#\# Writing Outline End \#\#\#\#

\#\#\#\# Completed Parts of the Paper Begin \#\#\#\#

\{paper\_already\}

\#\#\#\# Completed Parts of the Paper End \#\#\#\#

\#\#\#\# Experimental Settings Begin \#\#\#\#

\{experiments\_setting\}

\#\#\#\# Experimental Settings End \#\#\#\#

\#\#\#\# Experimental Results and Analysis Begin \#\#\#\#

\{experiment\_results\_and\_introduction\}

\#\#\#\# Experimental Results and Analysis End \#\#\#\#

\end{tcolorbox}

\begin{tcolorbox}[
    promptbox,
    title=Conclusion,
]

Based on the completed sections of the paper, write the conclusion section:
1. Ensure it follows the academic writing style of ICLR.

2. The conclusion should be one paragraph.

3. Content that requires LaTeX should be enclosed in \verb|```latex```| and use LaTeX syntax.

\#\#\#\# Completed paper section begin \#\#\#\#

\{paper\_already\}

\#\#\#\# Completed paper section end \#\#\#\#

\end{tcolorbox}

\begin{tcolorbox}[
    promptbox,
    title={Revise}
]

1. Revise the specified section of the paper according to the reviewer's suggestions.

2. Only revise the content and keep the style consistent with the previous one.

3. Enclose LaTeX content within triple backticks using \verb|```latex```|, and ensure all formatting follows LaTeX syntax.

\#\#\#\# Section {file} to be revised begin \#\#\#\#

\{paper\_now\}

\#\#\#\# Section {file} to be revised end \#\#\#\#

\#\#\#\# Reference for other sections begin \#\#\#\#

\{paper\_all\}

\#\#\#\# Reference for other sections end \#\#\#\#

\#\#\#\# Reviewer's comments begin \#\#\#\#

\{reviewer\_comments\_ex\}

\#\#\#\# Reviewer's comments end \#\#\#\#
\end{tcolorbox}

\subsection{Prompt for Review Agent}
\label{app:prompt_review}
All review-related prompts are consistent with \cite{jin2024agentreviewexploringpeer}

\section{Example}

\subsection{Example of LLM-authored Paper}
\label{app:example}
\begin{figure}[t]
    \centering
    \includegraphics[width=\textwidth,height=0.72\textheight,keepaspectratio]{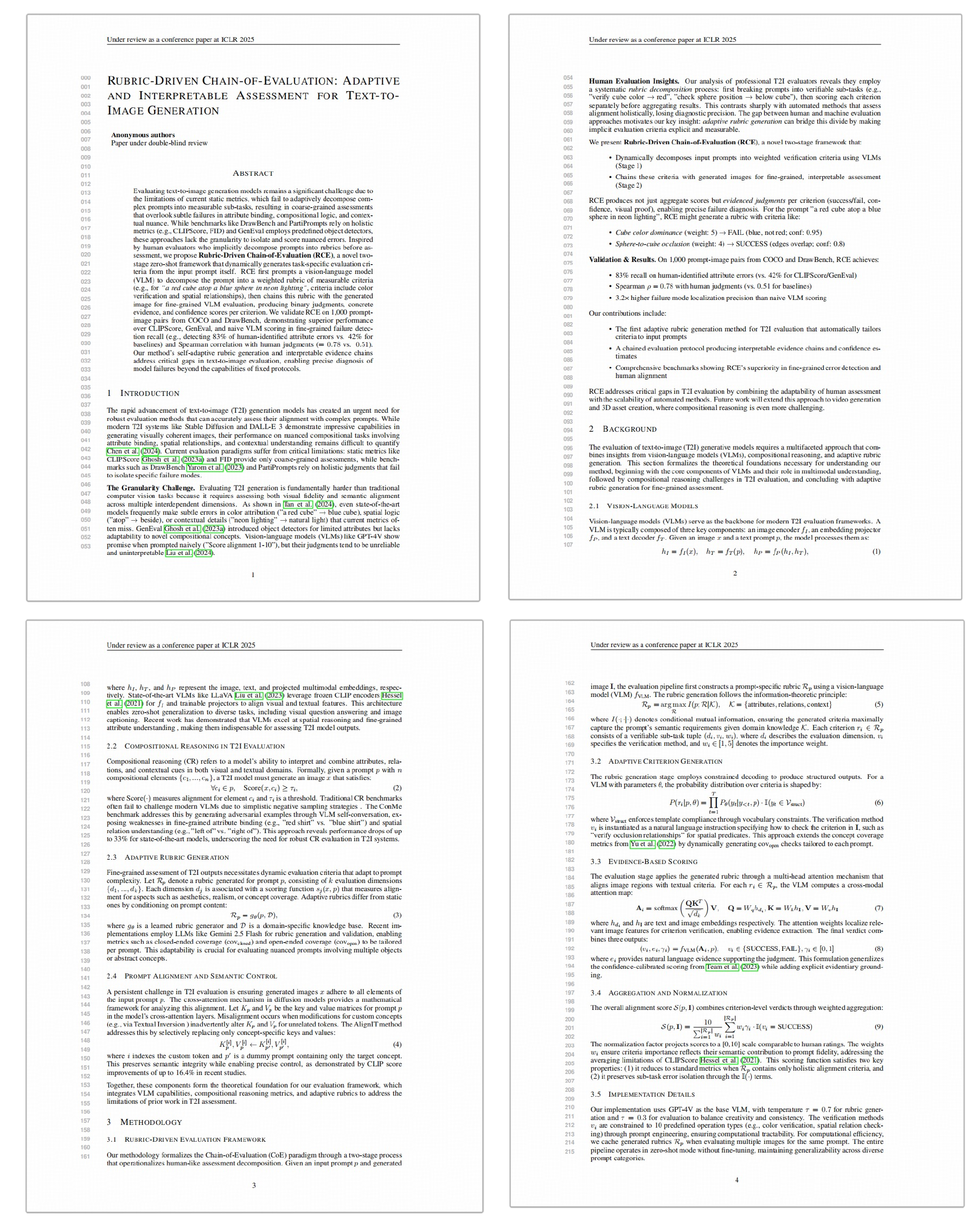}
    \caption{LLM-authored Paper Example}
    \label{fig:llm_author_expaple}
\end{figure}

One example of an LLM-authored paper is shown in Figure \ref{fig:llm_author_expaple}

\subsection{Example of LLM Review}
\label{app:llm_review}

\begin{tcolorbox}[
    reviewbox,
    title={Reviewer Assessment I}
]

\textbf{Reviewer 1}

Overall rating: 6\\\\Significance and novelty: The paper addresses the relatively unexplored topic of fairness in large language models (LLMs) when employed for tabular data prediction. This subject is significant given the extensive use of tabular data in high-stakes domains, and the potential social biases inherent in LLMs could lead to substantial risks in these applications.\\\\Reasons for acceptance:\\1. The paper tackles an underexplored area concerning the fairness of LLMs in tabular data prediction, which is crucial due to the wide application of tabular data in high-stakes domains.\\2. It provides empirical evidence showing the persistence of social biases in LLMs, which are inherited from pre-training data, significantly impacting fairness in predictions.\\3. The study explores different strategies to mitigate biases, such as in-context learning and fine-tuning, and compares their effectiveness against traditional models like Random Forests and Neural Networks.\\4. The findings contribute valuable insights into the limitations of current bias mitigation techniques for LLMs, suggesting further research and development of methods tailored to address inherent biases effectively.\\\\Reasons for rejection:\\1. **Methodological Concerns**\\   - The experimental design lacks detailed explanations on the criteria used for selecting few-shot examples and the mechanism of label flipping during in-context learning, which are crucial for replicability and understanding of the results.\\   - The finetuning process is briefly mentioned without a comprehensive description of parameters and settings, which are essential for evaluating the effectiveness of the approach.\\\\2. **Data and Results Interpretation**\\   - The paper does not sufficiently explore the implications of the observed biases in practical scenarios, missing an opportunity to connect experimental findings with real-world applications and consequences.\\   - There is a lack of discussion on the variability of results across different datasets, which could provide insights into dataset-specific challenges in bias mitigation.\\\\3. **Limited Comparative Analysis**\\   - The comparison with traditional models is somewhat superficial, lacking a deep dive into why LLMs perform differently and the specific characteristics of LLMs that might contribute to biased outcomes.\\   - There is insufficient analysis of how the proposed bias mitigation techniques perform across various datasets, limiting the generalizability of the findings.\\\\4. **Insufficient Exploration of Mitigation Strategies**\\   - The exploration of bias mitigation strategies such as data resampling is limited and lacks depth, particularly in contrasting their effectiveness with findings from traditional machine learning approaches.\\   - The paper does not propose novel techniques or strategies for bias mitigation beyond those already established, missing an opportunity to advance the field.\\\\Suggestions for improvement:\\1. Enhance the explanation and methodology behind the selection and implementation of few-shot examples and label-flipping techniques in in-context learning to ensure clarity and reproducibility.\\2. Provide a more detailed exploration of the practical implications of social biases observed in LLM predictions for tabular data, connecting experimental results with potential real-world applications and consequences.\\3. Conduct a deeper and more comparative analysis of LLMs against traditional models, focusing on specific characteristics that contribute to biased outcomes and exploring dataset-specific challenges in bias mitigation.\\4. Broaden the exploration of bias mitigation strategies, including the potential development of novel methods tailored to address inherent biases in LLMs, and provide a clear comparison of their effectiveness across various datasets.

\dashedline
      
\textbf{Reviewer 2}

Overall rating: 4\\\\Significance and novelty: The paper explores a timely and relevant issue, investigating the fairness of large language models (LLMs) when applied to tabular data, highlighting the transfer of social biases from training data to predictions. The novelty lies in focusing on fairness implications in high-stakes domains where such biases could have significant impacts. However, the contributions are primarily incremental, reiterating known biases in LLMs in a new context.\\\\Reasons for acceptance: \\1. The study addresses a critical issue of fairness in LLMs, particularly relevant for high-stakes applications using tabular data.\\2. It provides a comprehensive evaluation across multiple datasets, enhancing the robustness of findings regarding biases in LLM predictions.\\3. The investigation into zero-shot and few-shot learning scenarios adds depth to understanding how LLMs operate in different contexts.\\4. The paper highlights the limitations of current mitigation strategies such as in-context learning and data resampling, pushing the field to consider more advanced solutions.\\\\Reasons for rejection:\\1. **Lack of Depth in Analysis**\\   - The paper does not deeply explore the root causes of biases beyond stating that they are inherited from training data, missing opportunities to hypothesize or test mechanisms.\\   - There is little to no discussion on the potential societal impacts of these biases, limiting the exploration of the real-world significance.\\\\2. **Limited Novelty in Findings**\\   - The findings primarily confirm existing knowledge about biases in LLMs, adding little new insight into the nature or mitigation of these biases.\\   - The comparison with traditional ML models does not provide substantial new perspectives beyond what previous literature has discussed.\\\\3. **Methodological Concerns**\\   - The choice of datasets and protected attributes, while common, does not push the boundaries of fairness research into less explored but equally important areas.\\   - Label-flipping experiments are interesting but are presented without thorough analysis of why they have the observed effects, which could inform future mitigation strategies.\\\\4. **Clarity and Structure**\\   - The paper's structure is somewhat repetitive, particularly in the sections discussing experiments, which may obscure the key contributions.\\   - Some methodological details, such as the choice of specific hyperparameters or experimental setups, are relegated to appendices, which could affect reproducibility and understanding.\\\\Suggestions for improvement: \\1. Enhance the depth of analysis by exploring the underlying mechanisms of bias transfer from training data to predictions in LLMs, potentially through qualitative analyses or theoretical explorations.\\2. Expand on the societal implications of the findings, discussing potential impacts in high-stakes domains and suggesting actionable steps for stakeholders.\\3. Introduce more diverse datasets and protected attributes to explore the generalizability of findings across different contexts and biases, which would strengthen the paper's contributions.\\4. Improve clarity by restructuring the results section to avoid repetition and ensure that key findings and their implications are clearly and succinctly presented.,

\dashedline

\textbf{Reviewer 3:}

Overall rating: 5\\\\ Significance and novelty: The paper addresses an important yet underexplored topic: the fairness of large language models (LLMs) in making predictions on tabular data. Despite the growing use of LLMs in various tasks, their application in tabular data and the associated fairness implications have not been extensively studied. This work aims to fill this gap by investigating the sources of information LLMs rely on and the extent of bias present in their predictions. The novelty lies in examining the effectiveness of in-context learning, label-flipping, and fine-tuning as strategies to mitigate biases in LLMs, compared to traditional machine learning models.\\\\Reasons for acceptance:\\1. The paper tackles a significant and timely issue regarding the use of LLMs in high-stakes domains, where fairness is crucial.\\2. It provides a comparative analysis of LLMs and traditional models in tabular data tasks, contributing to a deeper understanding of biases in LLMs.\\3. The investigation into various bias mitigation strategies, such as in-context learning and fine-tuning, offers valuable insights into their effectiveness in enhancing fairness.\\4. The paper includes a comprehensive experimental setup and evaluation, testing on multiple datasets and employing several fairness metrics.\\\\Reasons for rejection:\\1. **Lack of sufficient novelty in methodological approach:**\\   - The use of in-context learning, fine-tuning, and label-flipping as bias mitigation strategies is not particularly novel, as these techniques have been explored in previous research.\\   - The paper provides limited innovation in terms of developing new methodologies or frameworks specifically designed for fairness improvement in LLMs for tabular prediction tasks.\\\\2. **Insufficient depth in analysis of results:**\\   - While the paper presents various experimental results, there is a lack of in-depth discussion regarding the underlying reasons for the observed biases and the varying effectiveness of different mitigation strategies.\\   - The paper does not offer a detailed exploration of why certain datasets or tasks exhibit more pronounced biases or why certain strategies are more successful than others.\\\\3. **Limited theoretical contribution:**\\   - The paper does not provide a strong theoretical foundation or model for understanding the inherent biases in LLMs, nor does it propose a new theoretical framework to guide future research in this area.\\   - The discussion primarily revolves around empirical findings without connecting them to broader theoretical insights in the field of AI fairness.\\\\4. **Inadequate exploration of potential solutions:**\\   - While the paper briefly touches upon advanced strategies needed to address biases, it falls short of proposing or testing novel solutions or enhancements beyond existing techniques.\\   - The paper should have included more exploration of new approaches or adaptations specifically tailored to address bias in LLM-based tabular data predictions.\\\\Suggestions for improvement:\\1. Introduce novel methodologies or frameworks specifically targeted at bias mitigation in LLMs for tabular tasks, enhancing the paper\u2019s originality and contribution to the field.\\2. Enhance the analysis and discussion of experimental results by providing deeper insights into the causes of biases and the differential effectiveness of mitigation strategies across various datasets and tasks.\\3. Develop a more robust theoretical contribution by proposing new models or frameworks to understand and address inherent biases in LLMs, linking empirical findings to theoretical insights in AI fairness.\\4. Explore and test innovative solutions or adaptations beyond current strategies to effectively reduce bias in LLMs, potentially incorporating interdisciplinary perspectives or new technologies.
\end{tcolorbox}

\begin{tcolorbox}[
    reviewbox,
    title={Author--Reviewer Discussion}
]

\textbf{Author:}

Response: Thank you for your insightful feedback. We appreciate the recognition of the significance and novelty of our work, as well as your constructive suggestions for improvement. Here are our responses to the concerns and suggestions you raised:\\\\1. **Methodological Concerns**: \\   - We acknowledge the need for a more detailed explanation of the criteria for selecting few-shot examples and the label-flipping mechanism. We will include a comprehensive description in the revised manuscript to enhance replicability and clarity.\\   - We will expand the section on finetuning to provide a thorough account of the parameters and settings used, ensuring that our approach is fully transparent and evaluable.\\\\2. **Data and Results Interpretation**:\\   - We agree that it's crucial to connect experimental findings with real-world applications. We will add a section discussing the practical implications of observed biases and potential consequences in high-stakes domains.\\   - To address the variability of results across different datasets, we will perform an additional analysis to explore dataset-specific challenges and include this discussion in the revised manuscript. \\\\3. **Limited Comparative Analysis**:\\   - We realize the importance of deeper comparative analysis with traditional models. We will enhance the discussion on specific characteristics of LLMs that might contribute to biased outcomes, and delve into why they perform differently from traditional models.\\   - We will conduct additional experiments to better understand the performance of bias mitigation techniques across various datasets, thus enhancing the generalizability of our findings.\\\\4. **Insufficient Exploration of Mitigation Strategies**:\\   - We will broaden our exploration of bias mitigation strategies, including a comparison with traditional machine learning findings, and aim to propose novel methods tailored specifically for LLMs.\\   - We appreciate the suggestion to advance the field by proposing new strategies and will look into innovative approaches to bias mitigation in future work.\\\\Your feedback is invaluable, and we are committed to addressing the points raised to improve the quality and impact of our paper. Thank you once again for your detailed review and suggestions.

\dashedline

\textbf{Author}

Response: Thank you for your review and insightful comments. We address each of your concerns below:\\\\1. **Lack of Depth in Analysis**: We agree that understanding the root causes of biases is crucial. While our paper emphasizes the inheritance of biases from training data, we acknowledge the need for a more in-depth exploration of the underlying mechanisms. Future work will focus on conducting qualitative analyses to hypothesize potential mechanisms of bias transfer and examining the societal structures that contribute to these biases. We will also reflect on the societal impact of these biases more thoroughly in our discussion section.\\\\2. **Limited Novelty in Findings**: We appreciate your perspective on the novelty of our findings. Our intent was to bridge the gap between known biases in LLMs and their implications in tabular data contexts. Acknowledging that this reiteration might not offer entirely new insights, we will focus on developing and testing novel mitigation strategies that address these biases more effectively.\\\\3. **Methodological Concerns**: We selected common datasets and protected attributes to provide a benchmark for fairness analysis. However, we understand your point about exploring less examined areas. We will expand our dataset selection to include a more diverse set with different protected attributes in future work. Regarding label-flipping, we recognize the need for deeper analysis, which we plan to address by investigating the semantic importance of flipped labels in future studies.\\\\4. **Clarity and Structure**: We apologize if the structure led to any confusion. We will ensure the results section is more concise and focused, avoiding unnecessary repetition to enhance clarity. Additionally, we will incorporate key methodological details, such as hyperparameters, in the main text to improve transparency and reproducibility.\\\\We appreciate your suggestions for improvement and will work to integrate them to strengthen the paper's contributions. Thank you again for your valuable feedback.

\dashedline

\textbf{Author:}

Response: Thank you for your detailed review and suggestions. We appreciate your positive remarks about the significance of our work and the comprehensive experimental setup. Below are our responses to the concerns raised:\\\\1. **Lack of sufficient novelty in methodological approach:** \\   - While in-context learning, fine-tuning, and label-flipping have been studied before, our work uniquely applies these strategies to the domain of tabular data with LLMs, which has not been extensively explored. We will emphasize this aspect more clearly in the revised version to highlight our contribution in adapting existing methods to a novel application domain.\\\\2. **Insufficient depth in analysis of results:**\\   - We acknowledge this concern and will include more detailed discussions on the underlying reasons for the observed biases and the differential effectiveness of bias mitigation strategies. We plan to conduct additional analysis to better understand why certain datasets exhibit more pronounced biases and include these insights in the revision.\\\\3. **Limited theoretical contribution:**\\   - We understand the need for a stronger theoretical foundation. In response, we will work on incorporating more theoretical insights into the paper, potentially by linking our empirical findings with existing theories in AI fairness. We will aim to propose preliminary theoretical models to better explain the biases observed.\\\\4. **Inadequate exploration of potential solutions:**\\   - We agree that exploring novel solutions is crucial. We plan to include proposed adaptations of existing techniques and potentially new strategies to mitigate bias more effectively. We will also consider interdisciplinary approaches that may offer fresh perspectives on this issue.\\\\We appreciate your constructive feedback and are committed to addressing these points to enhance the quality and impact of our paper. Thank you again for your valuable input.
    
\end{tcolorbox}

\begin{tcolorbox}[
    reviewbox,
    title={Reviewer Assessment II}
]

\textbf{Reviewer 1:}

Overall rating: 6\\\\Summary: The authors have provided thoughtful responses to most of the concerns raised, demonstrating a clear understanding of the feedback provided. They have acknowledged the need for enhanced methodological clarity, a deeper analysis of results, and a stronger theoretical contribution, and have outlined how they plan to address these aspects in future work or revisions of the paper. They also intend to include more diverse datasets and propose novel bias mitigation strategies, which would strengthen the study's contributions.\\\\However, the proposed updates are largely prospective, and the current work still lacks significant novelty and depth in certain areas. The findings, while important, primarily confirm existing knowledge and do not introduce groundbreaking methodologies or theoretical insights. Additionally, while the fairness analysis is comprehensive, there is still a lack of exploration into practical applications and real-world implications of the observed biases. The paper also remains limited in its exploration of innovative mitigation strategies and dataset diversity.\\\\Overall, the study provides a solid empirical foundation and addresses a critical and relevant problem, but there is room for improvement in terms of novelty, depth of analysis, and actionable contributions. The authors' plans to address these gaps in a revised manuscript are promising, hence the score is maintained at 6.

\dashedline
    
\textbf{Reviewer 2:}

Overall rating: 5\\\\Summary: The authors have provided a detailed response to the raised concerns, demonstrating a clear understanding of the feedback. They acknowledge the limitations in methodological clarity, depth of analysis, and exploration of novel mitigation strategies. They have outlined plans to address these issues in future work, including diversifying the datasets, improving explanations for existing methods like label-flipping, and proposing novel bias mitigation strategies. They also intend to explore the societal implications of biases observed in LLMs for tabular data tasks more thoroughly. \\\\However, while the authors' response suggests promising directions for future work, the rebuttal does not fully address the core limitations of the current paper. The study remains primarily incremental, mainly reiterating known biases in LLMs and their challenges in fairness metrics without introducing significant novelty. Methodological concerns, such as the lack of detailed explanations for key experimental setups (e.g., criteria for few-shot selection or label flipping), persist in the current work. Additionally, the comparative analysis with traditional models remains somewhat superficial, and the discussion on the generalizability of findings lacks depth. The absence of any innovative or advanced strategies for bias mitigation continues to limit the paper's potential contribution.\\\\While the authors have shown a willingness to improve upon these aspects, the current submission does not sufficiently address these gaps, leading to a reduction in the overall rating. The work offers valuable observations on biases in LLMs for tabular tasks but falls short of providing actionable contributions or significant advancements in methodology.

\dashedline

\textbf{Reviewer 3:}

Overall rating: 5\\\\Summary: The authors have provided detailed responses to the concerns raised, demonstrating an understanding of the feedback and a commitment to addressing these issues in future revisions. They acknowledge the need for enhanced methodological clarity and detail, particularly concerning the selection of few-shot examples and label-flipping techniques. The authors have also recognized the importance of exploring the societal implications of biases in LLMs and plan to connect experimental findings with real-world applications more thoroughly. Additionally, they aim to incorporate more diverse datasets and propose novel bias mitigation strategies tailored to LLMs for tabular tasks.\\\\However, while the authors' response outlines promising directions for future work, the current paper remains limited in terms of novelty and depth of analysis. The study primarily reconfirms known biases in LLMs without introducing significant methodological or theoretical advancements. The experimental design lacks detailed explanations for key aspects, and the comparative analysis with traditional models is somewhat superficial. There is also a lack of exploration into innovative bias mitigation strategies and the generalizability of findings across different contexts. These limitations continue to impact the paper's potential contribution to the field.\\\\Overall, the authors' planned improvements are promising, but the current submission does not sufficiently address these gaps, leading to a maintained score. The work offers important observations on biases in LLMs for tabular tasks, but it falls short of providing significant advancements or actionable contributions.

\end{tcolorbox}

\begin{tcolorbox}[
    reviewbox,
    title={Meta-Review Compilation}
]
\textbf{AC:}

Score: 5\\\\Summary: The paper examines the fairness of large language models (LLMs) when used for predictions on tabular data, a significant issue due to the wide application of tabular data in high-stakes domains. It identifies that LLMs inherit social biases from their training data, impacting fairness in predictions. The authors explore several bias mitigation strategies, such as in-context learning and fine-tuning, comparing their effectiveness with traditional models like Random Forests and Neural Networks. While these efforts are commendable, the paper primarily confirms existing biases in LLMs without introducing novel methodologies or deeper theoretical insights into mitigating these biases. The authors acknowledge the need for more diverse datasets and novel mitigation strategies and plan to incorporate these aspects in future work. Overall, the study presents valuable empirical observations but lacks substantial innovation and depth, particularly in exploring advanced bias mitigation strategies and real-world implications of the observed biases.,

\end{tcolorbox}

\begin{tcolorbox}[
    reviewbox,
    title={Final Decision}
]
Reviewer Assessment I stage Scores: 6, 4, 5\\
Reviewer Assessment II stage Scores: 6, 5, 5\\
Meta-Review Compilation stage Scores: 5\\
Average Score: (6 + 4 + 5 + 6 + 5 + 5 + 5 )/7 = 5.14

\end{tcolorbox}

\section{Detail}
\subsection{Negative Keywords}

As shown in Table \ref{tab:negative_keywords}, we provide a list of negative keywords used for the limitation-oriented topic analysis in the study.
The keyword set was constructed as a descriptive aid for analyzing limitation-oriented topics. We first collected candidate terms through vocabulary expansion around themes such as model limitations, risks, fairness, adversarial vulnerability, privacy and security concerns, failure modes, and mitigation. We then manually curated the list to remove overly broad or context-irrelevant terms. The keyword analysis is used as a broad quantitative check of topic patterns, rather than as the sole evidence for the paper's findings or as a causal test of review behavior.

\begin{table*}[h]
\centering
\renewcommand{\arraystretch}{1.2} 
\caption{List of negative keywords used for the limitation-oriented topic analysis in the study.}
\label{tab:negative_keywords}
\begin{tabular}{p{14cm}}
\toprule
\multicolumn{1}{c}{\textbf{Negative Keywords}} \\
\midrule
gap, failure, unclear, minimal, mismatch, hazard, challenging, inequalities, adversarially, defect, injustice, unable, amplify the exposure, harmful, performance drop, mitigations, inherent bias, vulnerable, attacks, falls short, detrimental, weakness, incapable, risk, threat, limitation, cautionary, stereotypes, misalignment, flaw, cannot reach, bias, drop, weaknesses, drawback, minimal impact, relative drop, inadequate, sparsity, pitfall, sensitivity, inability, shortcoming, biases, deep-rooted biases, erroneous, inconsistencies, implicit stereotypical, reject, cautionary tale, prejudice \\
\bottomrule
\end{tabular}
\end{table*}

\section{Additional Statistics}
\label{app:additional_stats}
\subsection{Human Paper Sampling Statistics}
\label{app:human_sampling_stats}

The human-authored papers used in the main simulation were sampled across ten research topics, with ten papers per topic, for a total of 100 papers. This sampling scheme was intended to cover a range of common machine learning and LLM-related directions rather than concentrating on a single area. Table~\ref{tab:human_sampling_stats} summarizes the final decision distribution and the corresponding OpenReview human-review scores. Accepted papers account for 41\% of the selected human-authored papers, indicating that the human baseline was not constructed by oversampling rejected papers.

\begin{table}[htbp]
    \centering
    \setlength{\tabcolsep}{4pt}
    \caption{Decision distribution and OpenReview human-review average scores for the selected human-authored papers.}
    \label{tab:human_sampling_stats}
    \begin{tabular}{lcc}
        \toprule
        \textbf{Decision} & \textbf{Count} & \textbf{Human Avg.} \\
        \midrule
        Accepted & 41 & 6.4577 \\
        Rejected & 59 & 4.6816 \\
        \midrule
        Total & 100 & -- \\
        \bottomrule
    \end{tabular}
\end{table}

\subsection{Human Annotation Details}
\label{app:human_annotation_details}

The human annotation study was designed as a targeted sanity check for cases where LLM reviewers most strongly favored LLM-authored papers, rather than as a full-scale expert-review replacement. We selected 15 matched human--LLM paper pairs with the largest LLM-favoring score gaps. Thirteen graduate-level or above CS annotators participated, including three co-authors and ten non-co-authors; seven annotators had prior reviewing experience. Each pair was independently evaluated by two annotators under ICLR-style review criteria, without revealing paper authorship.

To reduce obvious source or authorship cues, figures, appendices, references, and related-reference descriptions were removed for the annotation interface. This preprocessing was used only for the human annotation study, not for the simulation itself. It may affect the perceived completeness of some papers, especially human-authored ones, and should be considered when interpreting the annotation results. Inter-annotator agreement, measured by Cohen's $\kappa$, was approximately 0.52, indicating moderate agreement.

\subsection{Linguistic Statistics}
Linguistic statistics of human-authored papers and LLM-authored papers are shown in Table \ref{tab:app_descriptive_stats}.

\begin{table*}[htbp]
    \centering
    \caption{Linguistic Statistics of Human Papers and LLM Papers (Excluding Appendices)}
    \label{tab:app_descriptive_stats}
    \begin{tabular}{llcc}
        \toprule
        \textbf{Group} & \textbf{Metric} & \textbf{Human} & \textbf{LLM} \\
        \midrule
        \multirow{3}{*}{Text length} & paper length & 5548.10 & 4614.71 \\
        & sentence length & 11.8464 & 9.3222 \\
        & paragraph length & 51.1221 & 39.8206 \\
        \midrule
        \multirow{3}{*}{Vocabulary diversity} & 1-gram & 0.2598 & 0.4321 \\
        & 2-gram & 0.7069 & 0.8172 \\
        & 3-gram & 0.8844 & 0.9045 \\
        \midrule
        \multirow{3}{*}{Difficulty} & Flesch-Kincaid Grade Level & 11.4963 & 13.2567 \\
        & Dependency Distance & 4.2404 & 5.2659\\
        & Subclause Ratio & 0.0326 & 0.0233 \\
        \bottomrule
    \end{tabular}
\end{table*}

\begin{figure}[t]
    \centering
    \includegraphics[width=0.85\textwidth,height=0.72\textheight,keepaspectratio]{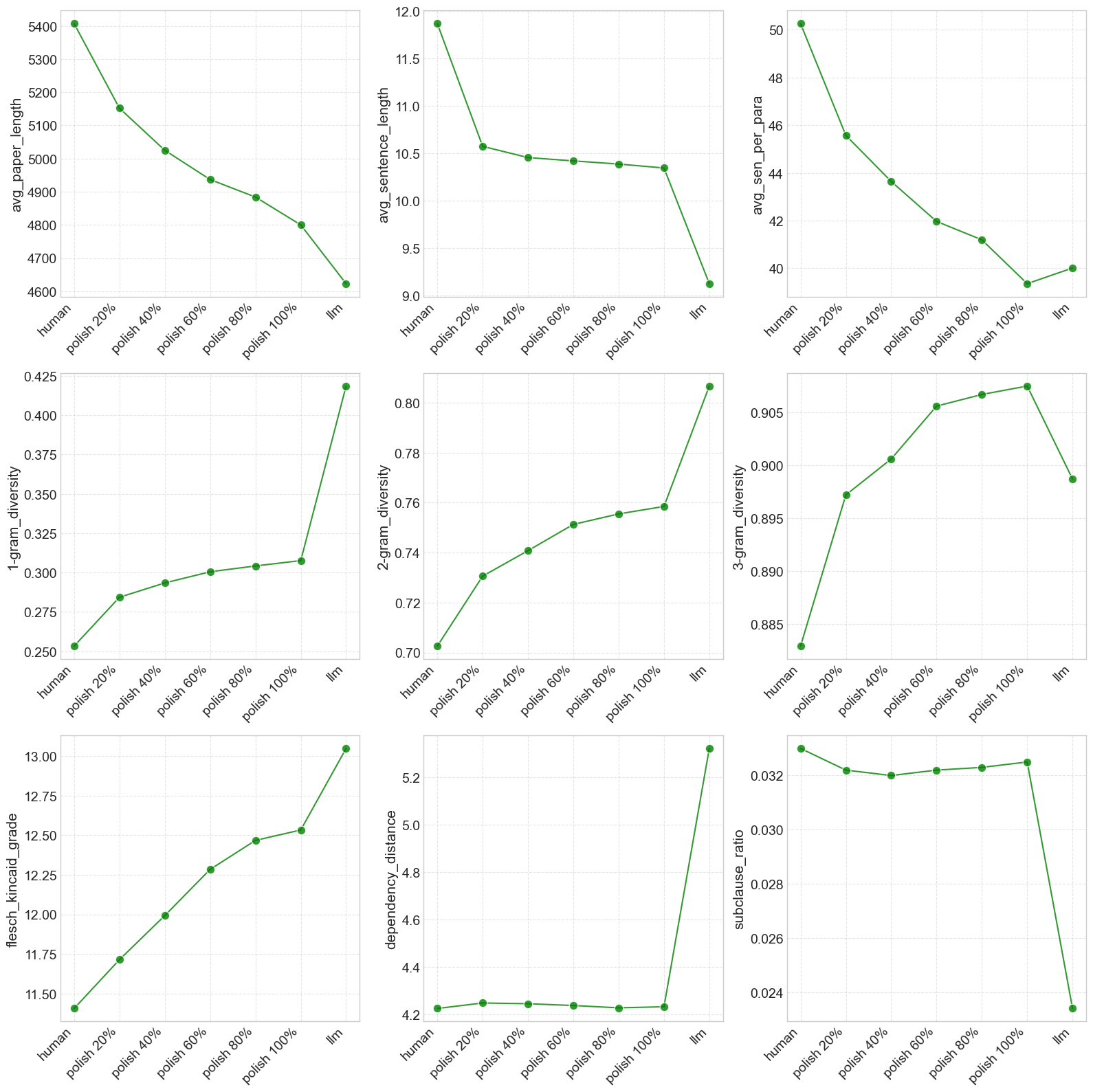}
    \caption{An analysis of the stylistic and linguistic features of LLM-generated text after varying levels of human polishing. We present a grid of nine plots, each corresponding to a distinct textual feature. Across all plots, the x-axis represents five document conditions: human-authored papers, polish 20\%, polish 40\%, polish 60\%, polish 80\%, and polish 100\%, as well as pure LLM-authored papers (llm). The y-axis shows the average value of each feature for the corresponding document condition. The features analyzed include length (avg\_paper\_length, avg\_sentence\_length, avg\_para\_length), lexical diversity (1-gram\_diversity, 2-gram\_diversity, 3-gram\_diversity), and complexity (flesch\_kincaid\_grade, dependency\_distance, subclause\_ratio).}
    \label{app:app_polish_ratio_sta}
\end{figure}

\begin{table*}[htbp]
    \centering
    \caption{Correlation between textual statistical features and target variable}
    \label{tab:app_correlation_text_feature}
    \begin{tabular}{llrr}
    \toprule
    \textbf{Metric} & \textbf{Feature} & \textbf{Correlation} & \textbf{P-value} \\
    \midrule
    \multirow{3}{*}{Length} 
        & paper\_length           & -0.1912 & 0.0070 \\
        & average\_sentence\_length & -0.1405 & 0.0483 \\
        & average\_sen\_per\_para   & -0.1412 & 0.0472 \\
    \midrule
    \multirow{3}{*}{Lexical diversity} 
        & 1-gram                  & 0.2795  & 0.0001 \\
        & 2-gram                  & 0.2386  & 0.0007 \\
        & 3-gram                  & 0.0745  & 0.2969 \\
    \midrule
    \multirow{3}{*}{Complexity} 
        & readability\_scores\_fkg & 0.1923  & 0.0067 \\
        & mean\_dependency\_distance & 0.1956  & 0.0057 \\
        & subclause\_ratio        & -0.2691 & 0.0001 \\
    \midrule
    \end{tabular}
    \end{table*}

\newpage

\section{Discussion}

\subsection{Additional Related Work}
\label{app:llm_reviewer_related}

\paragraph{LLMs as Reviewers}
Recent work has increasingly discussed potential reforms in the AI conference peer-review ecosystem. \citet{lie2024sentiment} proposes a sentiment-consolidation framework to improve meta-review generation. 
Building on concerns about review reliability, \citet{ryu2025reviewscoremisinformedpeerreview} introduces a criterion for detecting misinformed review points, such as questions already answered in the paper or critiques based on incorrect premises. Complementary to these technical approaches, several position papers explore broader structural reforms. 
\citet{anonymous2025position} argues against using LLMs for end-to-end review generation, instead recommending targeted uses (e.g., reproducibility checks, citation validation, ethics flagging) that reduce reviewer burden while preserving human expertise. 
Similarly, \citet{yang2025position} advocates for a more transparent, open, and regulated peer-review process, comparing fully open, partially open, and closed models. 
In parallel, \citet{kim2025position} calls for transforming the review pipeline from a one-way system into a bidirectional feedback loop, where authors evaluate review quality and reviewers receive formal accreditation through mechanisms such as two-stage review and systematic incentives, thereby supporting a sustainable and higher-quality peer-review framework.

\subsection{Justification for Using Predicted Results}

\paragraph{Constrained result setting}
To mitigate the possibility that LLM-generated papers benefit from overly optimistic synthetic results, we constrain their simulated numerical improvements to be no higher than those observed in the corresponding human-authored papers. Specifically, for each LLM-generated paper, reported improvements over baselines are capped by the improvement magnitudes observed in the matched human-authored paper under the same topic or keyword setting. This control reduces the chance that LLM-generated papers receive higher review scores simply because they report stronger experimental results, while preserving the scalability of the simulation.

\paragraph{Practical constraints and budget considerations}
(1) Implementing full experimental iterations would prevent the research process from being fully automated. (2) LLM coding capabilities are limited to less complex code, which would significantly restrict the scope of simulation topics. (3) Experimental iterations are time-consuming, making the paper generation process prohibitively long. Therefore, we believe it is acceptable to simplify the research agent's experimental execution, especially since this allows us to focus our effort and resources on exploring LLM reviewer risks rather than building a perfect research agent.

\subsection{Idea Consistency in Simulation}
\label{app:idea_consistency}
Idea generation is an important component of auto research systems. Therefore, if we replaced LLM-generated ideas with ideas extracted from human papers, the resulting LLM papers would not fully reflect the model's capabilities across the full research pipeline.
Additionally, this introduces an extra concern regarding whether the model can accurately extract the core ideas from human-authored papers. Inaccurate extraction could still lead to skewed results.
Thus, we preserve the idea generation step. We also believe that evaluating full-paper generation (including ideation) rather than restricting to writing style better aligns with real-world reviewing practices. In this context, topic-level consistency (i.e., ensuring that both human- and LLM-authored papers target the same research area or keywords) offers a reasonable level of control for paper comparison.

In addition, within the same topic, LLM-authored papers show significantly higher score distributions than human-authored papers. Paired comparisons of human and LLM papers under the same keywords further support LLM-authored paper superiority from different views. The comparison is conducted at the broader level of human-author versus LLM-author given the same specific domain, while keeping idea consistency is relatively fine-grained.

Moreover, we examined a setting in which the ideas remained unchanged while retaining the original results. As shown in Section~\ref{sec:pos_neg}, simply applying LLM-based polishing to human-authored papers led to a noticeable increase in average review scores, from 5.69 to 5.94.

Allowing the LLM to generate ideas led to another noteworthy observation: only human-authored papers ended up in an irreducibly rejected state, even after multiple rounds of LLM revision. We found that LLMs tend to generate more positive ideas, while some human-authored ideas are critical in nature. As a result, all LLM-authored papers were eventually accepted, whereas a subset of human-authored papers consistently received low scores.

\subsection{Rejected Human Papers}

We did not oversample rejected human-authored papers in the main human--LLM comparison. As shown in Table~\ref{tab:human_sampling_stats}, the selected human-authored set includes both accepted and rejected papers, with accepted papers accounting for 41\% of the sample. This indicates that the observed patterns are not driven by concentrating on rejected human submissions.

Additionally, Section~\ref{sec:human_check} reports a manual pairwise check for the largest LLM-favoring score gaps, which further rules out the interpretation that the effect is simply due to stronger LLM-authored papers.

\subsection{Automated Scientific Discovery}

\paragraph{Automated Scientific Discovery (ASD)} Prior work has explored LLMs' idea generation capabilities, which can be considered a form of automated scientific discovery. Large-scale human annotations indicate that LLM-generated ideas tend to be more novel \citep{si2024canllmsgenerate} but less actionable \citep{si2025ideationexecutiongapexecutionoutcomes} than human-generated ideas. While these studies provide valuable context, a fine-grained evaluation of ideas, such as assessing their reliability, is beyond the scope of our work.

\paragraph{Fabricated Scientific Discoveries and ASD Detection} The issue of fabricated scientific discoveries is not unique to LLMs; human-authored papers can also report fabricated results. Conversely, an LLM may generate speculative yet potentially valid scientific hypotheses. Existing hallucination detection methods could assist with ASD detection. However, unless a paper contains explicit contradictions or factual errors, determining whether a scientific discovery is fabricated, particularly at review time, is inherently challenging. Therefore, while ASD detection is important, it represents a long-standing problem that is conceptually distinct from the focus of our research.

\paragraph{Practical Scenario} With the increasing use of LLMs as reviewers (and researchers), human- and LLM-authored papers may compete directly, as some authors may even submit fully LLM-generated papers out of curiosity or irresponsibility. 
Among human-authored papers, some may receive substantial LLM-based polishing, while others receive minimal assistance; some emphasize strong positive results, whereas others highlight risks, limitations, or constraints. When such papers are evaluated by LLM reviewers, the observed score disparities could disadvantage certain groups of human authors.

\subsection{Human Review Noise}

\paragraph{Human Noise} In OpenReview, some reviewers are careless or outside their expertise. This noise makes perfect agreement unrealistic. However, in general, high-quality papers tend to be accepted and low-quality papers rejected, or the conference will lose its authority. Therefore, the human review data we collected can reasonably serve as a ground truth rather than ``human unreliability''. In addition, during our human-check stage, we invited responsible annotators to verify findings, confirming observed score disparities rather than suggesting that LLM-authored papers are superior.

\subsection{Agents Construction}

\paragraph{Research Agent}
We chose DeepSeek to serve as the backbone of our research pipeline, because its generated papers matching our expectations in terms of visual structure and content, and remains affordable for large-scale generation. Based on our preliminary investigation and testing, existing code models cannot run experiments and produce full research papers without human participation, and employing them introduces substantial time costs due to repeated trial-and-error. Consequently, we designed a dedicated pipeline, integrating LLM APIs and retrieval tools, to achieve automatic and scalable generation.

\paragraph{Review Agent}
The reviewer agent is a component of our simulation framework. Given that an effective review agent already exists, we believe introducing a new review agent and analyzing its risks offers limited value, and the risks we identify are rooted in the LLMs themselves. We report validation and additional reviewer-backbone checks in the main text.

\subsection{Review Scope}

Data and code are important components of the review process. However, based on both practical experience and official conference guidelines, submitted papers are not required to release data or code at review time, and most reviewers primarily evaluate the paper itself. Public statistics from ICLR 2024 indicate that only about 50\% of accepted papers released code (Poster: 881/1807; Spotlight: 183/367; Oral: 44/86). Moreover, even when data and code are provided, reviewers typically focus on the paper rather than verifying the code or datasets. Therefore, our evaluation protocol, which emphasizes the paper content rather than code or data, aligns with actual reviewing practices.

\end{document}